\documentclass[10pt,twocolumn,letterpaper]{article}
\usepackage{marvosym}
\usepackage{pdfpages}

\usepackage{cvpr}              

\definecolor{cvprblue}{rgb}{0.21,0.49,0.74}
\usepackage[pagebackref,breaklinks,colorlinks,citecolor=cvprblue]{hyperref}

\def\paperID{5932} 
\def\confName{CVPR}
\def\confYear{2024}

\title{Learning without Exact Guidance: Updating Large-scale High-resolution Land Cover Maps from Low-resolution Historical Labels}
\usepackage{multirow} 
\author{Zhuohong Li$^{1*}$, Wei He$^{1*}$, Jiepan Li$^1$, Fangxiao Lu$^1$, Hongyan Zhang$^{1, 2\dag}$\\
$^1$Wuhan University \hspace{+0.7em} $^2$China University of Geosciences\\
\tt\small \{ashelee, weihe1990, jiepanli, fangxiaolu\}@whu.edu.cn, zhanghongyan@cug.edu.cn
}

\begin{document}
\maketitle
\begin{abstract}
Large-scale high-resolution (HR) land-cover mapping is a vital task to survey the Earth's surface and resolve many challenges facing humanity. 
However, it is still a non-trivial task hindered by complex ground details, various landforms, and the scarcity of accurate training labels over a wide-span geographic area. In this paper, we propose an efficient, weakly supervised framework (Paraformer) to guide large-scale HR land-cover mapping with easy-access historical land-cover data of low resolution (LR). 
Specifically, existing land-cover mapping approaches reveal the dominance of CNNs in preserving local ground details but still suffer from insufficient global modeling in various landforms. Therefore, we design a parallel CNN-Transformer feature extractor in Paraformer, consisting of a downsampling-free CNN branch and a Transformer branch, to jointly capture local and global contextual information. Besides, facing the spatial mismatch of training data, a pseudo-label-assisted training (PLAT) module is adopted to reasonably refine LR labels for weakly supervised semantic segmentation of HR images. Experiments on two large-scale datasets demonstrate the superiority of Paraformer over other state-of-the-art methods for automatically updating HR land-cover maps from LR historical labels.  \renewcommand{\thefootnote}{}
\footnotetext{$^*$Indicates equal contribution. 
$^\dag$Corresponding author. The code and data are released at \url{https://github.com/LiZhuoHong/Paraformer}}
\end{abstract}    
\vspace{-1em}
\section{Introduction}
\label{sec:intro}

Land-cover mapping is a semantic segmentation task that gives each pixel of remote-sensing images a land-cover class such as "cropland" or "building" \cite{land-cover}. The land-cover data should be continuously updated since nature and human activities frequently change the landscape \cite{robinson2019large}. As sensors and satellites developed, massive high-resolution (HR) remote-sensing images ($\le$ 1 meter/pixel) could be easily obtained \cite{TONG2020111322}. Rapid large-scale HR land-cover mapping is even more critical to facilitate downstream applications as the up-to-date HR land-cover data can accurately describe the land surface \cite{li2022outcome,girard2021polygonal,zhang2022seamless}. However, the complex ground details reflected by HR images and various landforms over wide-span areas still challenge the periodic updating of large-scale HR land-cover maps \cite{li2022breaking}. 
\begin{figure}[]
{
    \begin{minipage}[b]{\hsize}
     \centering
    \includegraphics[width=\linewidth]{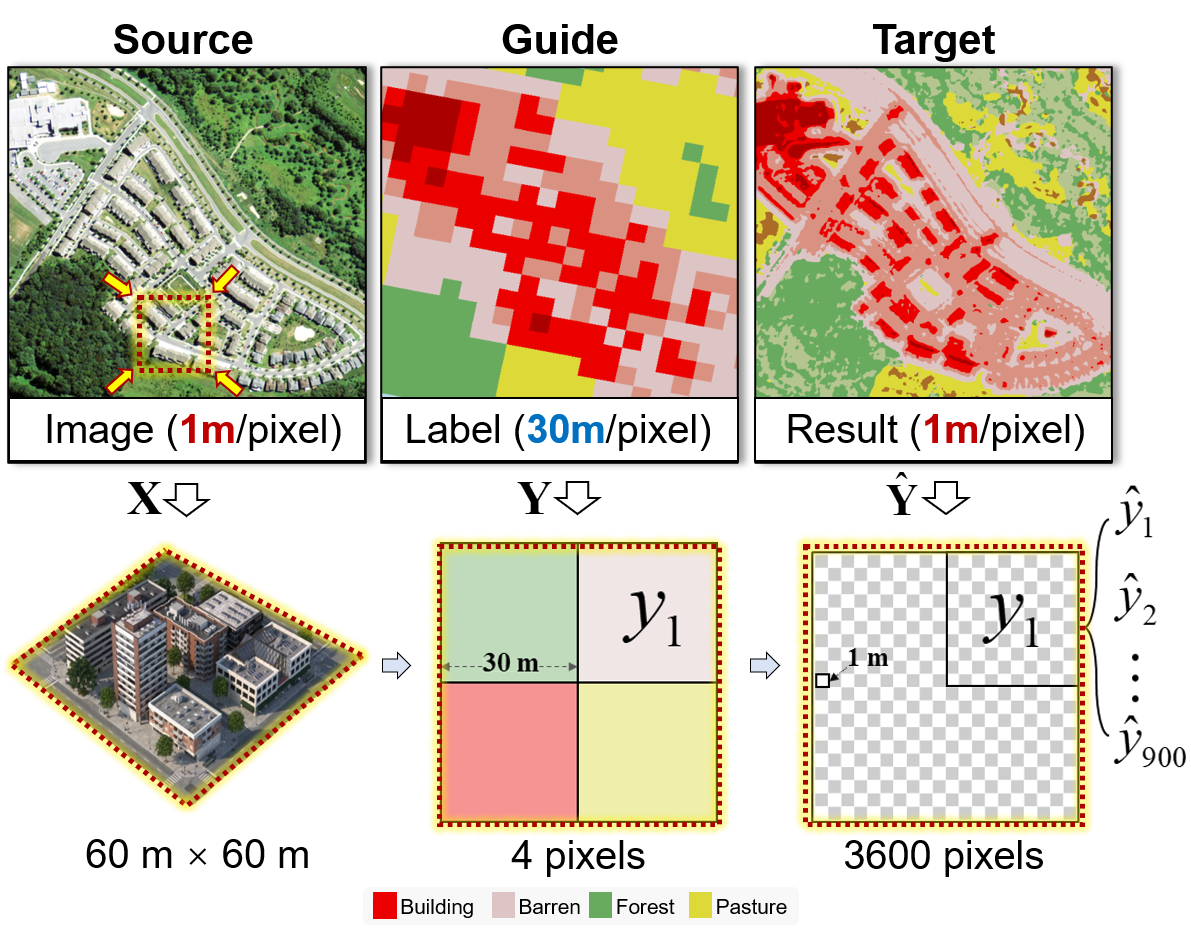}
    \end{minipage}
    } 
    \vspace{-2em}
\caption{ \footnotesize \rmfamily Illustration of resolution mismatched issue in using the HR remote-sensing image (\textbf{Source}) and LR historical labels (\textbf{Guide}) to generate HR land-cover results (\textbf{Target}).}
\label{intro}
\vspace{-1em}
\end{figure}

The advanced methods for HR land-cover mapping have been dominated by the convolutional neural network (CNN) for many years. Although CNN-based models can finely capture local details for semantic segmentation of HR images, the intrinsic locality of convolution operations still limits their implementation in various landforms across larger areas \cite{LUO2022105}.
Recently, Transformer has achieved tremendous success in semantic segmentation \cite{d2021convit, mehta2021mobilevit, cai2023efficientvit} and large-scale applications of Earth observation \cite{wang2022unetformer,sun2022multi,chen2021building}. It adopts multi-head self-attention mechanisms to model global contexts but struggles in the representation of local details due to the shortage of low-level features \cite{wang2022unetformer,chen2021transunet}.
\begin{figure}[!h]
\centering          
\includegraphics[width=0.65\linewidth]{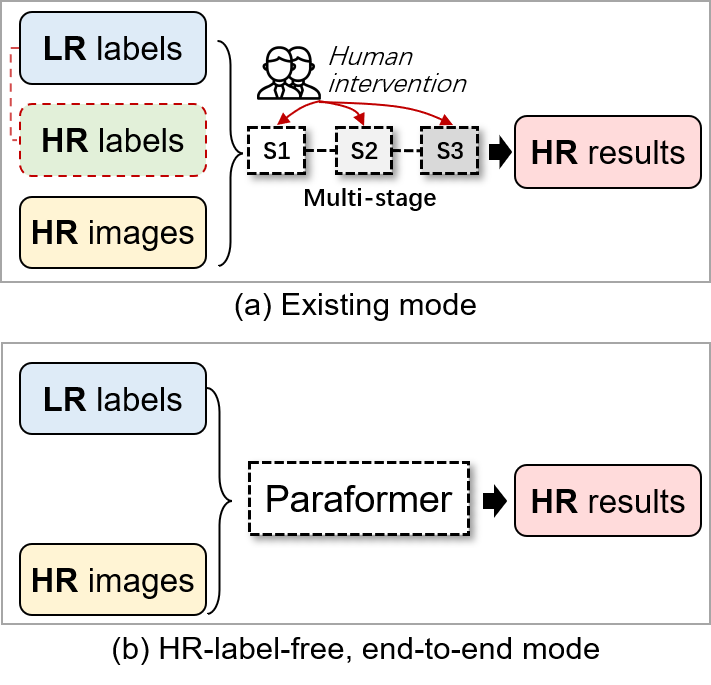}
\vspace{-1em}
\caption{ \footnotesize\rmfamily Two modes of large-scale HR land-cover mapping with LR labels. (a) Existing modes either reply on partial HR labels or require non-end-to-end training with human interventions. (b) \textbf{Paraformer} aims to form a mode that is HR-label-free and end-to-end trainable.} 
\vspace{-1em}
\label{diff modes}
\end{figure}
Besides, current methods with either CNN or Transformer structures generally rely on sufficient exact training labels by adopting a fully supervised strategy \cite{marcos2018land,gaetano2018two,sertel2022land}. However, creating accurate HR land-cover labels for large-scale geographic areas is extremely time-consuming and laborious \cite{robinson2019large,cao2022coarse}. 

Fortunately, many low-resolution (LR) land-cover data with large coverage have already emerged in the past decades \cite{zhang2021glc_fcs30,chen2019stable,van2021esa,karra2021global}. 
Utilizing these LR historical land-cover data as alternative guidance is a way to alleviate the scarcity of HR labels \cite{li2023sinolc}. 
Nevertheless, the unmatched training pairs of HR images and inexact LR labels posed a challenge for fully supervised methods. Moreover, due to the different applied scenarios, existing weakly supervised semantic segmentation methods for natural scenes (e.g., learning from bounding box or image-level labels) are not applicable in handling the challenge as well \cite{Dai_2015_ICCV, Zhou_2022_CVPR,Li_2022_CVPR,Lee_2022_CVPR}.

Distinctively, the incorrect samples of LR land-cover labels are brought by satellites in different spatial resolutions during Earth observation.
As shown in Figure \ref{intro}, the objects in a $60 m\times 60 m$ area can be clearly observed from the HR (1 m/pixel) image ${\bf{X}}$. However, in the LR (30 m/pixel) label ${\bf{Y}}$, the area is only labeled by four pixels. To produce the 1-m land-cover result ${\bf{\hat Y}}$, a labeled pixel ${y_1}$ needs to provide guiding information for 900 target pixels $\left\{ {{\hat y_{\rm{1}}},{\hat y_{\rm{2}}} \cdots {\hat y_{900}}} \right\}$, which raises a serious geospatial mismatch. How to reasonably exploit LR labels as the only guidance for semantic segmentation of large-scale HR satellite images is a particular problem shared in the fields of Earth observation and computer vision \cite{li2022breaking,robinson2019large,malkin2018label}. By summarizing the state-of-the-art methods of exploiting LR labels for large-scale HR land-cover mapping, there are still two main problems:
\begin{enumerate}
    \item \emph{For the wide-span application areas, existing feature extractors are difficult to jointly capture local details from HR images and model global contexts in various landforms at once \cite{li2023sinolc, yokoya20202020}.}

    \item \emph{For the mismatch of training pairs, existing pipelines, as shown in Figure \ref{diff modes} (a), either still rely on partial HR labels or require non-end-to-end optimization with human interventions
     \cite{chen2023novel,li2022outcome}.}  
    
\end{enumerate}

To resolve these problems, as shown in Figure \ref{diff modes} (b), we propose the Paraformer as an HR-label-free, end-to-end framework to guide large-scale HR land-cover mapping with LR land-cover labels. Specifically, Paraformer parallelly hybrids a downsampling-free CNN branch with a Transformer branch to jointly capture local and global contexts from the large-scale HR images and adopts a pseudo-label-assisted training (PLAT) module to dig up reliable information from LR labels for framework training.

The main contributions of this study are summarized as
follows: \textbf{(a)} We introduce an efficient, weakly supervised Paraformer to facilitate large-scale HR land-cover mapping by getting rid of the well-annotated HR labels and human interventions during framework training; \textbf{(b)} a downsampling-free CNN branch is parallelly hybridized with a Transformer branch to capture features with both high spatial resolution and deep-level representation. The structure aims to globally adapt large-scale, various landforms and locally preserve HR ground details; \textbf{(c)} the PLAT module iteratively intersects primal predictions and LR labels to constantly refine labeled samples for guiding the framework training. It provides a concise way to update large-scale HR land-cover maps from LR historical data. 

\vspace{-0.5em}
\begin{figure*}[t]
\centering
\includegraphics[width=0.75\linewidth]{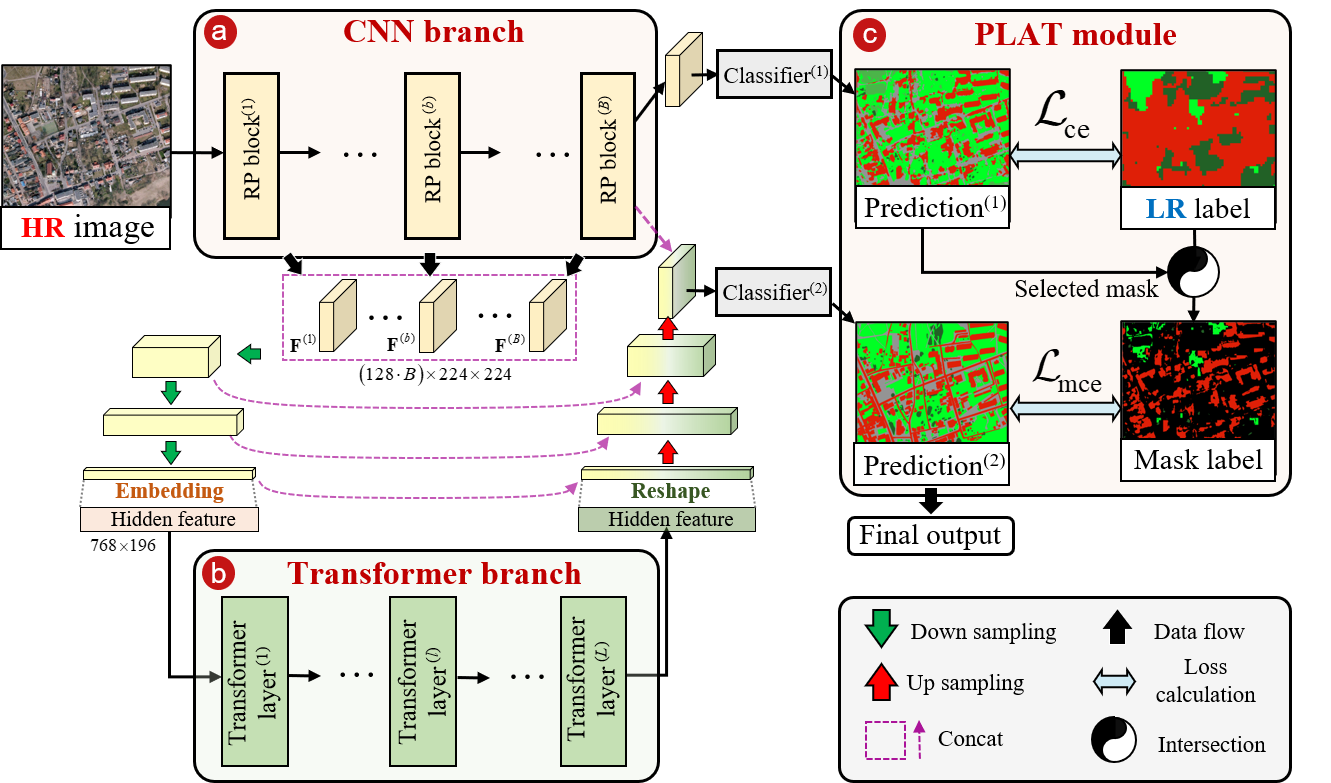}
\caption{ \footnotesize\rmfamily Overall workflow of Paraformer. The framework only takes the HR images and LR labels as training input and includes three components: (a) CNN-based resolution-preserving branch, (b) Transformer-based global-modeling branch, and (c) Pseudo-Label-Assisted Training (PLAT) module. } 
\label{framework}
\vspace{-1em}
\end{figure*}
\section{Related Work}
\noindent
\textbf{Land-cover mapping approach:} 
In the early stage, pixel-to-pixel classification methods, such as decision tree \cite{friedl1997decision}, random forest \cite{chan2008evaluation}, and support vector machine \cite{shi2015support}, were popular in the land-cover mapping of multi-spectral LR images. However, these methods generally ignore contextual information and have fragmented results in HR cases, as optical HR images contain abundant spatial details but limited spectral features \cite{li2023sinolc}. With the development of data-driven semantic segmentation, many CNN-based models were widely used in land-cover mapping of HR images \cite{robinson2019large,xu2022luojia,xie2021super}. Besides, as an alternative architecture, Transformer shows great power in capturing global contexts with sequence-to-sequence modeling \cite{Liu_2021_ICCV, chen2021transunet, WANG2022196} and demonstrates outstanding performance in many large-scale applications of Earth observation, such as building extraction \cite{sun2022multi, 10418227}, road detection \cite{chen2021building}, and land-object classification \cite{wang2022novel}. 
Besides, many works developed new ways by saving labor to produce finer labels with the Segment Anything Model (SAM) \cite{wu2023samgeo,osco2023segment}.
However, sufficient exact training labels are the foundation for large-scale applications of both CNN- and Transformer-based methods. 
The scarcity of HR labels still impedes these fully supervised approaches from large-scale HR land-cover mapping.\\
\textbf{Land-cover labeled data:} Creating large-scale HR labels via manual and semi-manual annotations is extremely time-consuming
and expensive \cite{pengra2015global,dong2021high}. Therefore, exiting HR land-cover data is generally limited to small scales. E.g., the LoveDA dataset contains 0.3-m land-cover data, covering 536.15 $km^2$ of China \cite{wang2021loveda}. The Agri-vision dataset contained 0.1-m labeled data, covering 560 $km^2$ of the USA \cite{chiu2020agriculture}. In the contract, the LR land-cover data generally has a larger coverage. E.g., the United States Geological Survey cyclically updates 30-m land-cover data covering the whole USA \cite{wickham2021thematic}.  The European Space Agency (ESA) has updated an annual 10-m global land-cover data since 2020 \cite{van2021esa}. These LR data can be seen as an alternative label source for guiding large-scale HR land-cover mapping. However, massive inexactly labeled samples still hinder them from being practicable.\\
\textbf{Strategies for LR historical label mining:}
To alleviate the scarcity of accurate labels in large-scale HR land-cover mapping, many studies have made efforts to mine reliable information from LR labels. E.g., a label super-resolution network was designed to constrain the inexact parts of LR labels by using the statistical distribution inferred from HR labels \cite{malkin2018label, robinson2019large}. A multi-stage framework, named WESUP, was built for 10-m land-cover mapping with 30-m labels \cite{chen2023novel}. In WESUP, multi-models were trained to refine clean samples from LR labels. Similarly, the winner approach of the 2021 IEEE GRSS Data Fusion Contest (DFC) deployed a shallow CNN to refine the 30-m labels, and then multi-model were trained with pseudo-labels to create the 1-m land-cover map of Maryland, USA \cite{li2022outcome}. Moreover, a low-to-high network (L2HNet) was proposed to select confident parts of LR labels via weakly supervised loss functions \cite{li2022breaking}. To produce 1-m land-cover maps across China with available 10-m labels, seven L2HNets were selectively trained to adapt wide-span geographic areas \cite{li2023sinolc}. 

Different from these approaches that either still rely on partial HR labels or require human interventions, Pafaromer is designed as an HR-label-free end-to-end framework to facilitate large-scale HR land-cover mapping.
\section{Methodology}

To jointly capture local and global contexts and reasonably exploit LR labels for large-scale HR land-cover mapping, Paraformer combines parallel CNN and Transformer branches with a PLAT module. In this section, the three components are introduced sequentially.
\subsection{CNN-based resolution-preserving branch}
As a basic feature extractor of Paraformer and also the main structure of previous L2HNet V1 \cite{li2022breaking}, the CNN branch is designed to capture local contexts from HR images and preserve the spatial details by preventing feature downsampling. As shown in Figure \ref{framework} (a), the CNN branch is constructed by five serially connected resolution-preserving (RP) blocks. Each RP block contains parallel convolution layers with the sizes of $1\times1$, $3\times3$, and $5\times5$, whose steps are set to 1 for feature size maintaining. Partly similar to the inception module \cite{Szegedy_2015_CVPR}, the channel numbers of different scales' layers in each block are inversely proportional to their kernel sizes, which are set to 128, 64, and 32. Based on the setting, the RP blocks can capture features with a proper receptive field instead of downsampling the feature maps with a deep encoder-decoder pattern. The serial blocks aim at sufficiently preserving the spatial resolution of features by using the majority of $1\times1$ kernels. The 3$\times$3 and 5$\times$5 kernels capture necessary surrounding information. Furthermore, the multi-scale feature maps are concatenated and reduced to 128 channels for branch lightening. Besides, a shortcut connection is adopted between blocks for residual learning and detail preserving.

\begin{figure}[]
{
    \begin{minipage}[b]{\hsize}
     \centering
    \includegraphics[width=\linewidth]{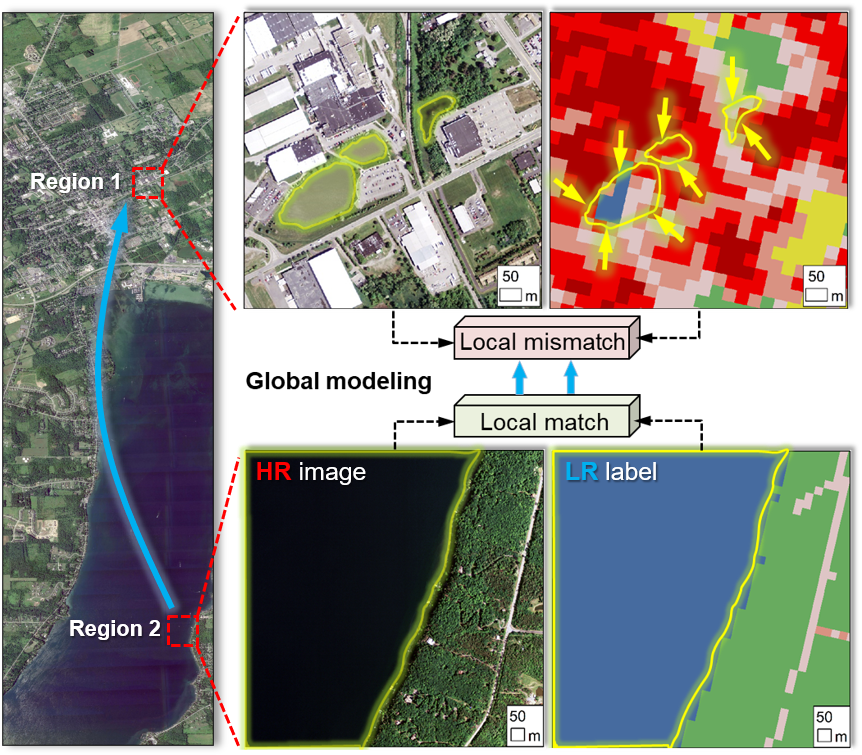}
    \end{minipage}
    } 
    \vspace{-2em}
\caption{ \footnotesize\footnotesize \rmfamily Example of the local mismatch/match in two regions. The edge of water is marked with \textcolor{yellow}{\textbf{yellow}} boundaries. Region 1 shows dispersed lakes around urban areas with unmatched annotation. Region 2 shows a large-scale river with matched annotation.}
\label{ViT}
\vspace{-1.2em}
\end{figure}
\subsection{Transformer-based global-modeling branch}
The ground objects with the same land-cover class may have distinctive attributes in HR images and are differently annotated in LR labels. Figure \ref{ViT} shows typical cases of lakes and rivers located in different areas. By considering that the CNN branch with intrinsic locality hinders the adaptation of various landforms over large-scale areas, we further hybrid the CNN branch with a Transformer branch which aims at capturing global contexts and building long-range support among dispersed geographic areas. As shown in Figure \ref{framework} (b), the Transformer branch contains 12 transformer layers. Each layer includes layer normalization, multi-head self-attention, and multi-layer perception. The feature maps extracted by each RP block are concatenated and inputted to the Transformer branch. Specifically, the extracted features from the CNN branch are downsampled and embedded in a hidden feature layer. And then the Transformer branch encodes the dense feature patches to capture global contexts. Subsequently, the encoded features are constantly upsampled to the size of HR images and classified to the final results. During the upsampling process, the outputted features of each stage are concatenated with the pre-encoded features, which bring massive local contextual information to the final feature maps. 


\subsection{Pseudo-Label-Assisted Training module}
To reasonably guide the large-scale HR land-cover mapping with weak LR labels, as shown in Figure \ref{framework} (c), a weakly supervised PLAT module is adopted to optimize the framework training. The PLAT module aims to screen out uncertain samples and dig up reliable information from the LR labels. Specifically, the two parts of the PLAT module are explained as follows.
For the CNN branch, we use classifier$^{(1)}$, which is constructed by $3 \times 3$ convolution layers, to generate the primal prediction$^{(1)}$ based on the extracted HR feature maps. Then we calculate the Cross-Entropy (CE) loss between prediction$^{(1)}$, represented as $\bf{\hat{Y}'}$, and the LR label, represented as $\bf{Y}$. Formally, by regarding $H$, $W$, and $L$ as the height, weight, and land-cover class of the patch, the CE loss of the CNN branch is written as:
\begin{equation} 
\small
{{\cal L}_{{\rm{ce}}}}({\bf{Y}},{\bf{\hat{Y}'}}) = \frac{{\sum\nolimits_{i = 0}^W {\sum\nolimits_{j = 0}^H {\left[ {\sum\nolimits_{l = 1}^L {y_{ij}^{(l)}\log (\hat{y}'^{(l)}_{ij})} } \right]} } }}{{H \times W}}.
\label{lce}
\end{equation}

\begin{figure*}[t]
\centering
\includegraphics[width=0.98\linewidth]{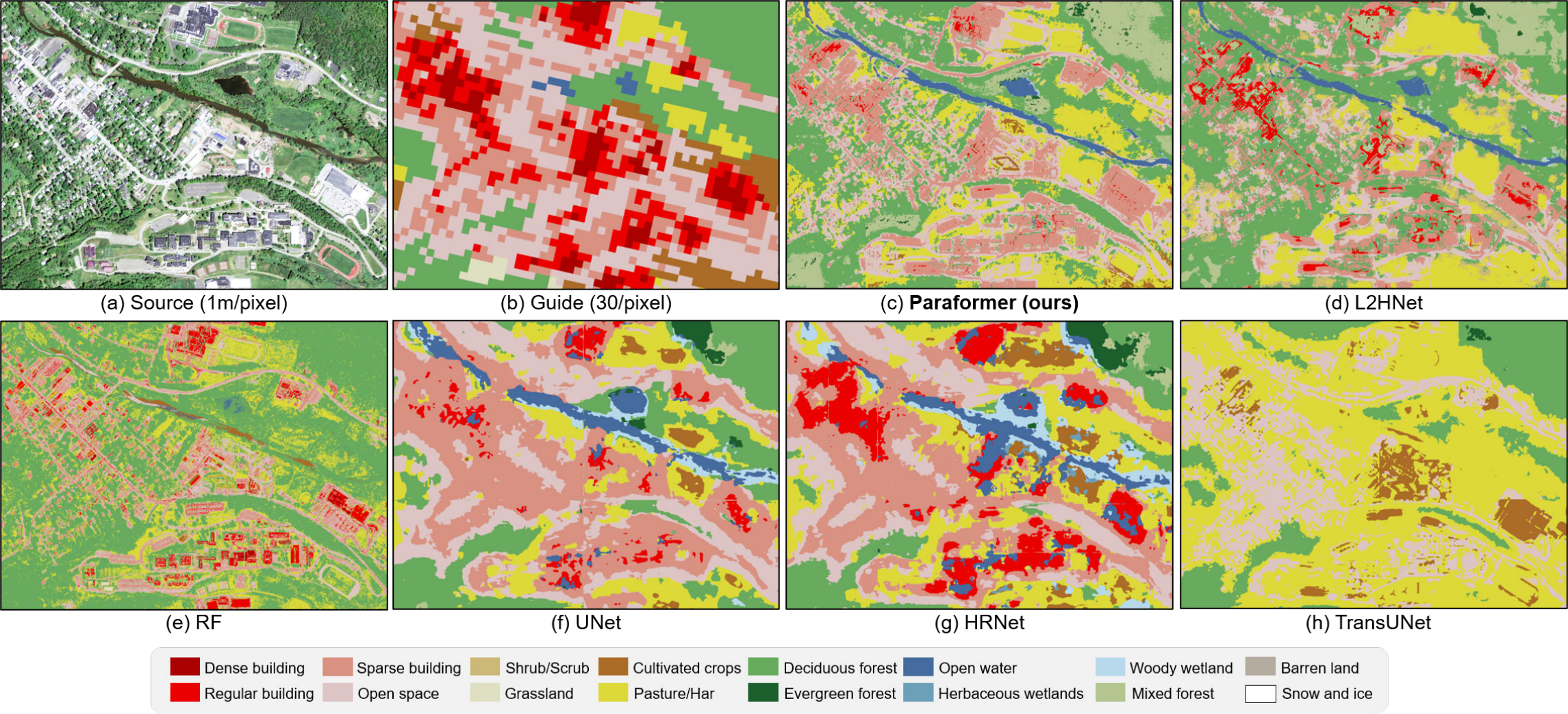}
\vspace{-1em}
\caption{ \footnotesize\rmfamily Demonstration of the training data and visual comparisons of the \textbf{Paraformer} and other typical methods on the Chesapeake Bay dataset with 16 classes. (a) HR image. (b) LR label. (c) land-cover mapping result of Parafomer. (d–h) land-cover mapping results of five typical methods.} 
\label{chesapeake qualitative result}
\vspace{-0.5em}
\end{figure*}
\begin{figure*}[t]
\centering
\includegraphics[width=0.98\linewidth]{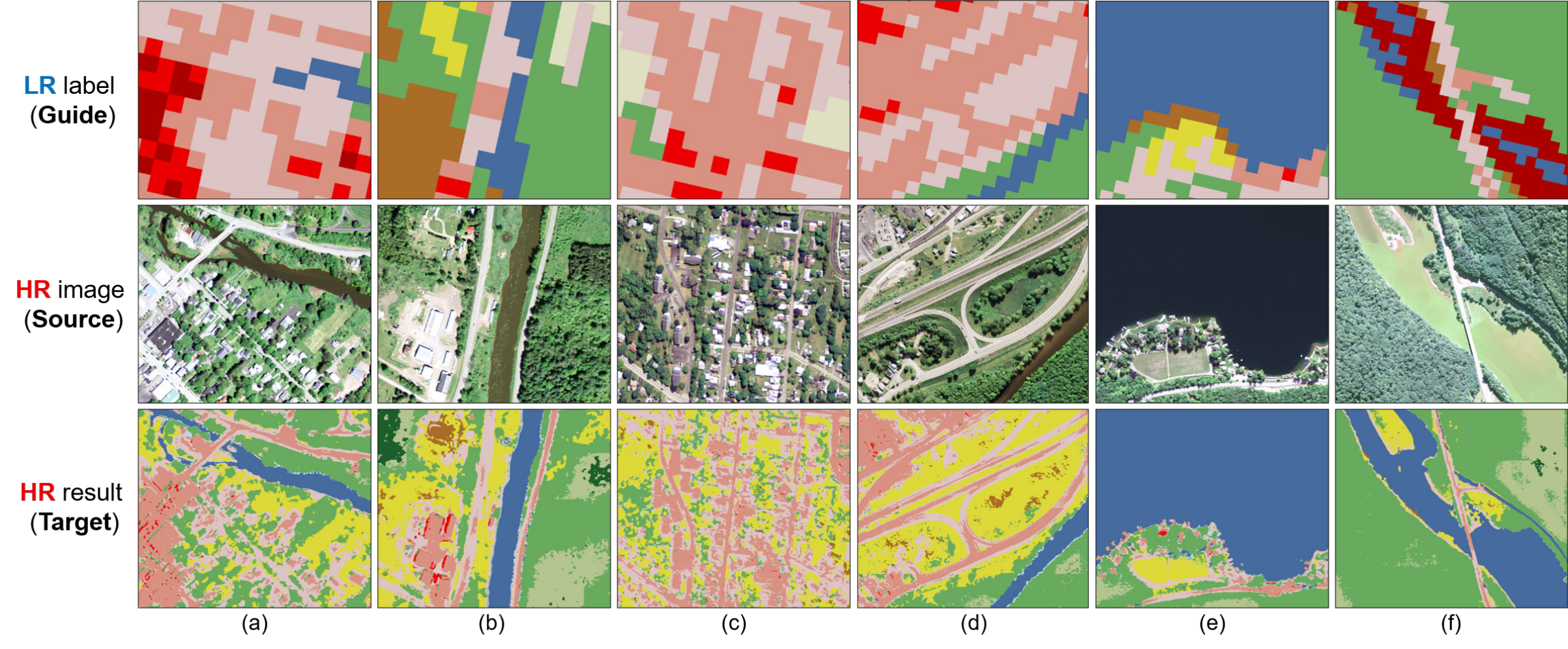}
\vspace{-1.5em}
\caption{ \footnotesize\rmfamily Six typical areas with finer observation scale on the Chesapeake Bay dataset. The first row shows the LR labels \textbf{(Guide)}. The second row shows the HR images \textbf{(Source)}. Third row shows the HR results \textbf{(Target)} produced by \textbf{Paraformer}.} 
\vspace{-1em}
\label{chesapeake finer result}

\end{figure*}
\begin{figure*}[!h]
\centering
\includegraphics[width=0.98\linewidth]{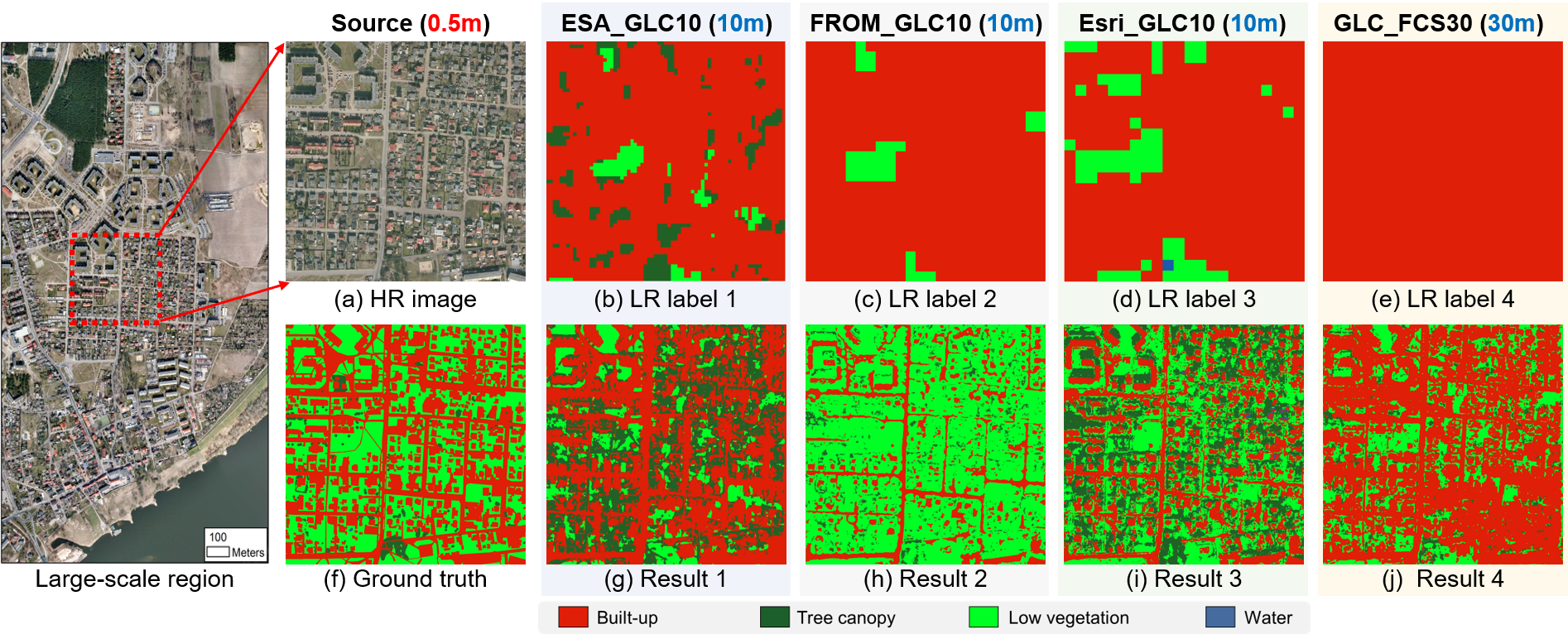}
\vspace{-1em}
\caption{ \footnotesize\rmfamily Visual results of \textbf{Paraformer} in the Poland dataset. The demonstration area is one of the training pieces sampled from large-scale training regions. (a–e) the training pairs of HR images (0.5 m/pixel) and four types of LR labels including ESA\underline{ }GLC (10 m/pixel), FROM\underline{ }GLC (10 m/pixel), Esri\underline{ }GLC (10 m/pixel), and GLC\underline{ }FCS30 (30 m/pixel). (f–g) the ground truth (0.5 m/pixel) and the mapping results of Paraformer with different LR labels.} 
\label{poland qualitative result}
\end{figure*}
\begin{table*}[t]
\centering
\footnotesize
\begin{tabular}{clccccccc}
\hline
\multirow{2}*{Resolution gap}  & \multirow{2}*{Method}  & \multicolumn{7}{c}{mIoU ($\%$) of six states in the Chesapeake Bay watershed}   \\
                         &                          & Delaware     & New York     & Maryland     & Pennsylvania     & Virginia     & West Virginia     & \textbf{Average}   \\ \hline
\multirow{15}*{$30\times$}   & \textbf{Paraformer}               & \centering\textbf{65.57} & \textbf{71.43} & \textbf{70.20} & \textbf{60.04} & \underline{68.01} & 52.62 & \textbf{64.65} \\
                         & L2HNet \cite{li2022breaking}                   & \underline{61.77} & \underline{68.12} & \underline{65.24} & 58.52 & \textbf{69.39} & \textbf{55.43} & \underline{63.08} \\
                         & TransUNet \cite{chen2021transunet}                & 53.15 & 60.53 & 60.42 & 51.08 & 66.21 & 47.52 & 56.49 \\
                         &ConViT \cite{d2021convit} &55.26&60.71&61.58&53.94&59.80&49.11&56.73\\
                          &CoAtNet \cite{dai2021coatnet}&56.89 &62.83&61.25    &53.57    &65.67    &51.34    &58.59     \\
                        &MobileViT\cite{mehta2021mobilevit}&58.03     &61.32     &61.84     &55.53    & 57.04   &48.64    &57.07      \\
                    &EfficientViT\cite{cai2023efficientvit}&53.72&61.28&59.48&51.38&57.34&48.76&55.33\\
                          &UNetFormer\cite{wang2022unetformer}& 58.85     &65.11     & 61.34    & \underline{59.10}   &60.84    &47.20    &58.74      \\
                        &DC-Swin\cite{wang2022novel}&59.65     &65.99     & 58.60    &58.06    & 64.11   & 48.15   &59.09      \\ 
                         & UNet \cite{ronneberger2015u}                     & 54.16 & 58.79 & 56.42 & 53.21 & 57.34 & 46.11 & 54.34 \\
                         & HRNet \cite{wang2020deep}                    & 52.11 & 56.21 & 50.76 & 50.03 & 57.48 & 45.42 & 52.00 \\
                         & LinkNet \cite{chaurasia2017linknet}                  & 58.27 & 62.05 & 52.96 & 52.11 & 48.71 & 48.93 & 53.84 \\
                         & SkipFCN \cite{li2021change}                  & 60.97 & 64.83 & 59.44 & 55.37 & 64.72 & \underline{54.66} & 60.00 \\
                         & SSDA \cite{tu2021high}                     & 57.91 & 61.54 & 54.85 & 51.71 & 57.71 & 47.15 & 55.15 \\
                         
                         & RF \cite{chan2008evaluation}                       & 59.35 & 55.03 & 55.26 & 51.07 & 52.29 & 54.36 & 54.56 \\ \hline                  \end{tabular}
\vspace{-1em}
\caption{\footnotesize The quantitative comparison of the Paraformer and other methods on six states of the Chesapeake Bay watershed. All methods were trained with the 1-m images and 30-m labels. The mIoU (\%) of different methods was calculated between their results and the 1-m ground truth.}
\label{Chesapeake results}
\end{table*}
\begin{table*}[!h]
\centering
\footnotesize
\begin{tabular}{clccccccccc}
\hline
\multirow{3}{*}{\begin{tabular}[c]{@{}c@{}}Max gap\end{tabular}} & \multirow{3}{*}{LR label} & \multicolumn{9}{c}{mIoU (\%) of different methods}                                                                     \\
 &                            &   \begin{tabular}[c]{@{}c@{}}\textbf{Paraformer}\\ \textbf{(ours)}\end{tabular}     & \begin{tabular}[c]{@{}c@{}}L2HNet\\ \cite{li2022breaking}\end{tabular} & 
 \begin{tabular}[c]{@{}c@{}}TransUNet\\  \cite{chen2021transunet}\end{tabular} 
 &
    \begin{tabular}[c]{@{}c@{}}ConViT\\  \cite{d2021convit} \end{tabular} 
 & 
    \begin{tabular}[c]{@{}c@{}}MobileViT\\  \cite{mehta2021mobilevit}\end{tabular} 
 &  
     \begin{tabular}[c]{@{}c@{}}DC-Swin\\  
     \cite{wang2022novel}\end{tabular} 
 & 
   \begin{tabular}[c]{@{}c@{}}HRNet\\  \cite{wang2020deep}\end{tabular} 
 & 
   \begin{tabular}[c]{@{}c@{}}SkipFCN\\  \cite{li2021change}\end{tabular} 
 & 
      \begin{tabular}[c]{@{}c@{}}RF\\  
     \cite{chan2008evaluation} \end{tabular} 
 \\ \hline
\multirow{3}{*}{$40\times$}                                                      & FROM\_GLC10 \cite{chen2019stable}                & \textbf{56.57} & \underline{50.15} & 38.44    & 39.36 & 41.03 & 43.56  & 43.66    & 27.14 & 21.48 \\
& ESA\_GLC10 \cite{van2021esa}                 & \textbf{55.19} & \underline{52.13} & 35.58    & 36.09 & 38.42 & 40.05  & 49.81    & 28.34 & 26.97 \\
& Esri\_GLC10 \cite{karra2021global}                & \textbf{55.07} & \underline{50.78} & 37.79    & 38.78 & 38.50 & 39.91  & 46.65    & 28.18 & 19.36 \\
$120\times$                                                                       & GLC\_FCS30 \cite{zhang2021glc_fcs30}                & \textbf{49.39} & \underline{43.62} & 26.20    & 29.16 & 29.57 & 30.14  & 41.46    & 23.67 & 17.02 \\ \hline    
\end{tabular}
\vspace{-1em}
\caption{\footnotesize The quantitative comparison on the Poland dataset. The mIoU (\%) of the Paraformer and other methods that trained with three types of 10-m labels (i.e., FROM\_GLC10, ESA\_GLC10, and Esri\_GLC10) and one type of 30-m label (i.e., GLC\_FCS30) are demonstrated.}
\vspace{-1em}
\label{Poland dataset}
\end{table*}

As the final output of the framework, prediction$^{(2)}$ is classified from the concatenated feature maps of CNN and Transformer branches, which is represented as $\bf{\hat{Y}''}$. During each training iteration, we take the simple but effective \textbf{intersection} of  prediction$^{(1)}$ and LR label to generate mask labels. Specifically, the inconsistent samples in mask labels are set as void values to remove them from loss calculations. Moreover, since predictions of the CNN branch contain HR textual information that is highly consistent with the images, the mask labels also outline fine edges and retain stable labeled samples.
Finally, the proposed Mask-Cross-Entropy (MCE) loss is calculated between prediction$^{(2)}$ and mask labels. Formally, the MCE loss is written as:
\begin{equation} 
\small
{{\cal L}_{{\rm{mce}}}}({\bf{M}}\cdot{\bf{Y}},{\bf{\hat{Y}''}}) = \frac{{\sum\nolimits_{i = 0}^W {\sum\nolimits_{j = 0}^H {\left[ {\sum\nolimits_{l = 1}^L {y_{ij}^{(l)}{m_{ij}}\log (\hat{y}''^{(l)}_{ij})} } \right]} } }}{{{\rm{Sum}}({\bf{M}}(i,j) = 1)}}.
\label{lmce}
\end{equation}
In Eqs. \ref{lmce}, ${\bf{M}}$ is the \textbf{intersected} mask with the size of $H \times W$. ${m_{ij}}, i \in \left[ {0,H} \right],j \in \left[ {0,W} \right]$ is the element of ${\bf{M}}(i,j)$ which can be simply represented as: 
\begin{equation} 
\small
{m_{ij}} = \left\{ \begin{array}{l}
\left. 1 \right|{Y_{ij}} = {{Y'}_{ij}}\\
\left. 0 \right|{Y_{ij}} \ne {{Y'}_{ij}}.
\end{array} \right.
\label{Mij}
\end{equation}
The total loss of the Paraformer is the combination of two branches' losses, which is written as:
\begin{equation} 
\small
{{\cal L}_{{\rm{total}}}} = {{\cal L}_{{\rm{ce}}}} + {{\cal L}_{{\rm{mce}}}}.
\label{Mij}
\end{equation}

\vspace{-1em}
\section{Experiments}

\subsection{Study areas and using data}
\vspace{-0.5em}
To comprehensively evaluate Paraformer on various landforms and different LR labels, the experiments are conducted on two large-scale datasets.\\
\textbf{The Chesapeake Bay dataset }is sampled from the largest estuary in the USA  and organized into 732 non-overlapping tiles, where each tile has a size of 6000 × 7500 pixels \cite{robinson2019large}. The specific data includes:
\begin{enumerate}

\item \emph{The HR images (1 m/pixel)} are from the U.S. Department of Agriculture’s National Agriculture Imagery Program (NAIP). The images contained four bands of red, green, blue, and near-infrared \cite{maxwell2017land}.

\item \emph{The LR historical labels (30 m/pixel)} are from the USGS's National Land Cover Database (NLCD) \cite{wickham2021thematic}, including 16 land-cover classes. 

\item \emph{The ground truths (1 m/pixel)} are from the Chesapeake Bay Conservancy Land Cover (CCLC) project.
\end{enumerate}
\textbf{The Poland dataset }contains 14 provinces of Poland and is organized into 403 non-overlapping tiles, where each tile has a size of 1024 × 1024 pixels. The specific data includes:
\begin{enumerate}

\item \emph{The HR images (0.25m and 0.5 m/pixel)} are from the LandCover.ai \cite{boguszewski2021landcover} dataset. The images contained three bands of red, green, and blue.

\item \emph{The LR historical labels} are collected from three types of 10-m land-cover data and one type of 30-m data, which are named FROM\underline{ }GLC10 \cite{chen2019stable}, ESA\underline{ }GLC10 \cite{van2021esa}, ESRI\underline{ }GLC10 \cite{karra2021global}, and GLC\underline{ }FCS30 \cite{zhang2021glc_fcs30}.

\item \emph{The HR ground truths} are from the OpenEarthMap \cite{xia2023openearthmap} dataset with seven land-cover classes.

\end{enumerate}
\begin{figure*}[t]
\centering
\includegraphics[width=0.9\linewidth]{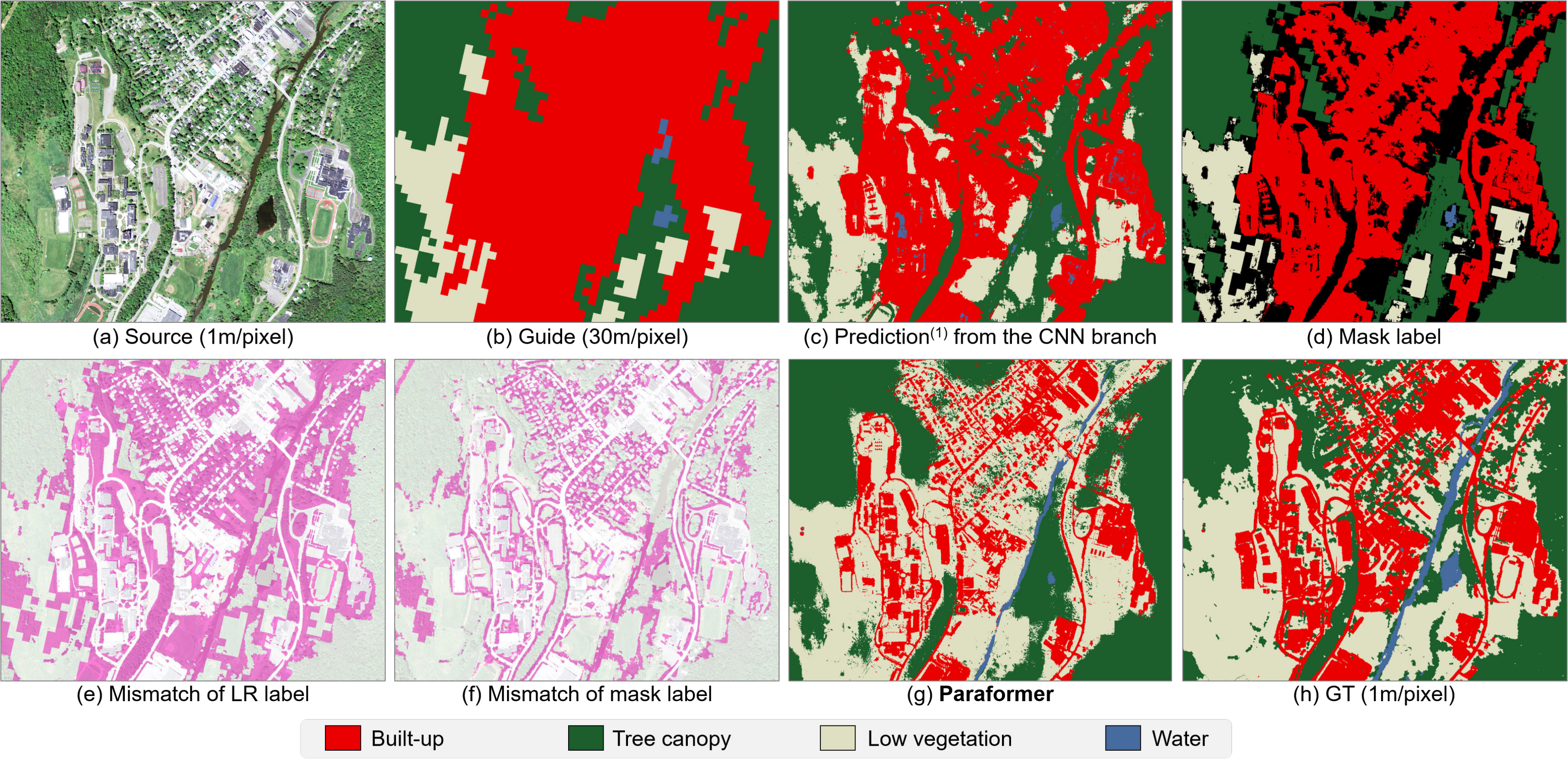}
\vspace{-0.5em}
\caption{ \footnotesize\rmfamily Example of training data and different outputs of Paraformer sampled from the Chesapeake Bay dataset with four unified classes. (a) HR images. (b) LR labels. (c) the primal prediction from the CNN branch. (d) Mask label, as the intersection parts of (b) and (c). The \textbf{black areas} are set to void without supervised information. (e–f) the incorrect samples (with \textcolor[rgb]{0.9,0.5,0.8}{\textbf{pink}} color) of LR label and mask label. (g) the final results of Paraformer. (h) HR ground truth.} 
\label{ablation1}
\vspace{-0.7em}
\end{figure*}

\subsection{Implementation Detail and Metrics}
\vspace{-0.5em}
In the experiments, all methods only take LR land-cover data as training labels. Paraformer is trained by the AdamW optimizer with a patch size of 224$\times$224 and batch size of 8. The learning rate is set to 0.01 and would decrease by 10\% when the loss stopped dropping over eight epochs. The metric of mean intersection over union (mIoU) is calculated between the results and the HR ground truths after their land-cover classes are unified into four base classes. 
The compared methods include: Random Forest (RF) is a pixel-to-pixel method widely used in large-scale land-cover mapping \cite{chan2008evaluation}. TransUNet \cite{chen2021transunet}, ConViT \cite{d2021convit}, CoAtNet \cite{dai2021coatnet}, MobileViT \cite{mehta2021mobilevit}, and EfficientViT \cite{cai2023efficientvit} are CNN-Transformer hybrid methods for semantic segmentation. UNetformer \cite{wang2022unetformer} and DC-Swin \cite{wang2022novel} are dedicated CNN-Transformer methods for remote-sensing images. UNet \cite{ronneberger2015u}, HRNet \cite{wang2020deep}, and LinkNet \cite{chaurasia2017linknet} are typical CNN-based semantic segmentation methods which are widely adopted in HR land-cover mapping \cite{robinson2019large,xu2022luojia,xie2021super}. SkipFCN \cite{li2021change} and SSDA \cite{tu2021high} are shallow CNN-based methods for updating 1-m land-cover change maps from 30-m labels, which won first and second place in the 2021 IEEE GRSS DFC  \cite{li2022outcome}. L2HNet is a state-of-the-art method designed for weakly supervised land-cover mapping \cite{li2022breaking}.
\subsection{Comparison Results}
\vspace{-0.5em}
\textbf{Comparison on the Chesapeake Bay dataset:} Table \ref{Chesapeake results} and Figure \ref{chesapeake qualitative result} show the comparisons on the Chesapeake Bay dataset. From the quantitative results, Paraformer shows superiority in the states of Delaware, New York, Maryland, and Pennsylvania. The L2HNet shows better results in Virginia and West Virginia. On average, Paraformer has the most accurate HR land-cover mapping results over the entire area, with a mIoU of 64.65\%. As shown in Figure \ref{chesapeake qualitative result} (c), the visual result of Paraformer is more consistent with the HR image compared with other methods. Unlike the fully supervised semantic segmentation task, the unmatched training pairs can cause serious misguidedness during the model training. E.g., as the rough results shown in Figure \ref{chesapeake qualitative result} (f) and (g), UNet and HRNet over-downsample the features and encourage results to fit LR labels instead of being consistent with the HR images. Furthermore, quantitative results reveal that UNet, LinkNet, and HRNet have insufficient performance, with mIoU of 54.34\%, 53.84\%, and 52.00\%. Although the compared CNN-Transformer methods (e.g., TransUNet) combine local and global contextual information, the structure does not focus on preserving the feature resolution or dealing with the geospatial mismatch. As a result, TransUNet shows a weak performance in visual results, shown in Figure \ref{chesapeake qualitative result} (h), and has a mIoU of 56.49\%. Furthermore, SkipFCN, SSDA, and RF use small receptive fields or pixel-to-pixel strategies to extract features with fine land details. However, due to the lack of deep-level feature representation and global contextual information, SkipFCN, SSDA, and RF obtain a mIoU of 59.99\%, 55.15\%, and 54.56\%, respectively. As an example shown in Figure \ref{chesapeake qualitative result} (e), RF finely predicts ground details but incorrectly classifies rivers, lakes, and pastures.
\begin{table*}[t]
\footnotesize
\centering
\begin{tabular}{lccccccccc}
\hline
               \multirow{2}*{Ablation method}              & \multicolumn{7}{c}{mIoU (\%) of six states in the Chesapeake Bay watershed}                                                              \\
                                            & Delaware              & New Your              & Maryland              & Pennsylvania              & Virginia              & West Virginia              &\multicolumn{1}{l|}{\textbf{Average}}   &  Params & FLOPs          \\ \hline
Paraformer          & \textbf{65.57} & \textbf{71.43} & \textbf{70.20} & \textbf{60.04} & \textbf{68.01} & \textbf{52.62} & \multicolumn{1}{l|}{\textbf{64.65}} & 109.4M & 141.3G\\
                        Sole CNN branch     & 59.57          & 67.87          & 64.30          & 53.86          & 65.26          & 50.01          & \multicolumn{1}{l|}{60.15} & 4.5M & 56.1G         \\
                        Sole Transformer branch     & 53.15          & 60.53          & 60.42          & 51.08          & 66.22          & 47.52          & \multicolumn{1}{l|}{56.49}    & 96.9M & 83.3G       \\
                        Hybrid without PLAT & \underline{62.69}    & \underline{70.39}    & \underline{67.15}    & \underline{58.33}    & \underline{67.47}    & \underline{50.83}    & \multicolumn{1}{l|}{\underline{62.81}}    & 109.4M & 141.3G \\ \hline

\end{tabular}
\vspace{-0.5em}
\caption{\footnotesize The ablation results of the Paraformer on six states of the Chesapeake Bay watershed. The sole CNN branch, sole Transformer branch, and Hybrid without PLAT aim to investigate the contribution of the CNN branch, Transformer branch, and PLAT module, respectively.}
\label{Ablation study}
\vspace{-1em}
\end{table*}
To further demonstrate the effect of Paraformer on different landscapes, we sample six typical areas in Figure \ref{chesapeake finer result}. The visual results indicate that the complex ground details among various landforms of HR land-cover maps can be well updated from the LR historical land-cover labels.\vspace{+0.5em}
\\
\textbf{Comparison on the Poland dataset:} In the experiments with the Poland dataset, all methods were used to produce 0.25/0.5-m land-cover maps of 14 provinces of Poland by exploiting four LR labels separately. These LR labels include 10-m FROM\underline{ }GLC10, ESA\underline{ }GLC10, Esri\underline{ }GLC10, and 30-m GLC\underline{ }FCS30. As shown in Table \ref{Poland dataset}, Paraformer is compared with eight representative methods (i.e., weakly supervised, CNN-Transformer, CNN-based, pixel-to-pixel approaches) in a more extreme geospatial mismatch. 
Compared with the state-of-the-art method, the Paraformer has an increase in mIoU of 6.42\%, 3.06\%, and 4.29\% in exploiting 10-m labels. By resolving 30-m labels with a max resolution gap of 120 $\times$, Paraformer has a mIoU of 49.39\% with an increase of 5.77\% compared with L2HNet. The typical CNN-based method has an average mIoU of 46.71\% among the 10-m cases and 41.46\% in the 30-m case. Skip\underline{ }FCN and RF have the lowest mIoU among all methods, which shows the difficulty of dealing with extremely unmatched situations. Moreover, the quantitative results of Paraformer shown in the four cases reveal that the proposed framework obtains stable results from different LR labels. 
Figure \ref{poland qualitative result} shows the visual results of Paraformer among four cases. With the parallel CNN-Transformer structure and PLAT module, Paraformer is able to refine the clear ground details (e.g., vegetation and roads) even if they are roughly labeled in local areas. In general, Paraformer shows the potential to robustly update large-scale HR land-cover maps from available LR historical labels.

\vspace{-0.5em}
\begin{figure}[]
\centering
\includegraphics[width=\linewidth]{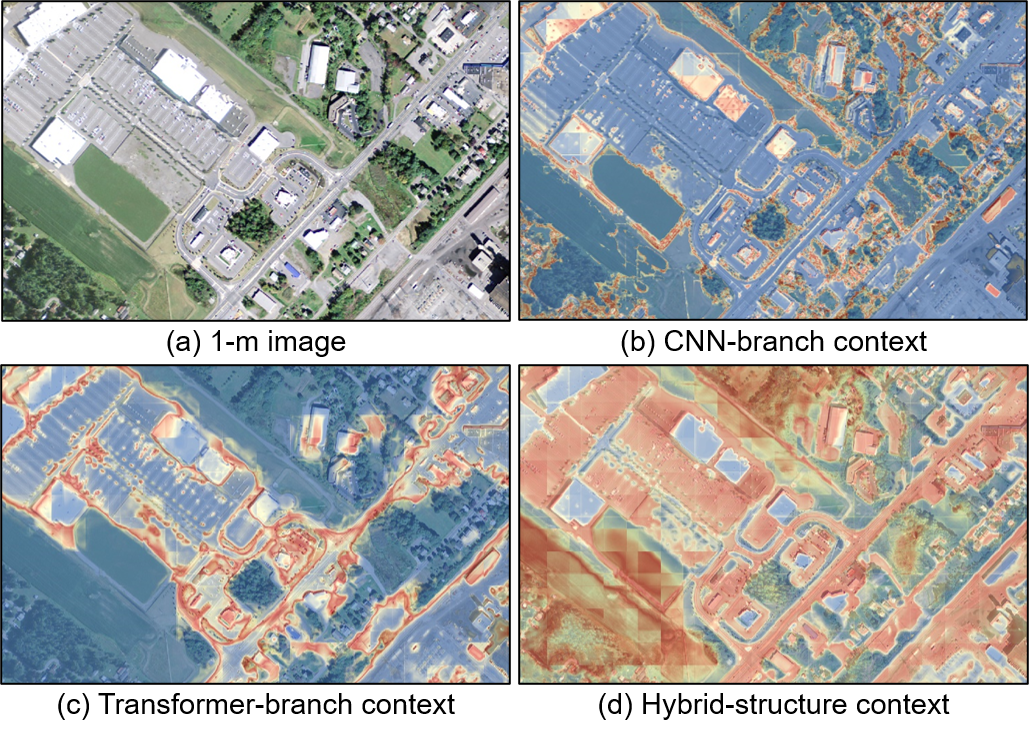}
\vspace{-2em}
\caption{ \footnotesize\rmfamily Demonstration of the extracted contexts from the ablation methods. (a) the original HR image. (b) the contexts extracted by the sole CNN branch. (c) the contexts extracted by the sole Transformer branch . (d) the contexts extracted by the CNN-Transformer hybrid backbone.} 
\label{ablation 2}
\vspace{-0.5em}
\end{figure}

\subsection{Ablation experiments}

In this section, ablation experiments were conducted on the Chesapeake Bay dataset to evaluate different components of Paraformer. Each ablation in Table \ref{Ablation study} is explained as follows: (1) the sole CNN branch is dependently trained by calculating CE loss with LR labels; (2) the sole Transformer branch embeds HR images instead of features from the CNN branch and calculates CE loss with LR labels; (3) the hybrid structure without PLAT directly calculates CE loss with the LR labels. 

By ablating the PLAT module, the results obtained an average mIoU of 62.81\%, which indicates a 1.84\% decrease compared with the 64.65\% of Paraformer. By ablating the CNN and Transformer branches, the results of the sole CNN branch obtained a mIoU of 60.15\% and had a 4.5\% decrease. Results of the sole Transformer branch obtained the lowest mIoU of 56.49\% and had the most obvious decrease (8.16\%). Figure \ref{ablation1} shows different outputs of Paraformer, where the inexact LR labels are gradually refined during framework training. The final result shown in Figure \ref{ablation1} (g) indicates both fine ground details and accurate land-cover patterns that are consistent with the ground truth. Moreover, Figure \ref{ablation 2} shows the visualized contexts captured by the CNN branch, Transformer branch, and hybrid structure. Figure \ref{ablation 2} (b) indicates that the CNN branch mostly focuses on capturing local details (e.g., the edges of roads, single houses, and shrubs). Figure \ref{ablation 2} (c) indicates that the Transformer branch captures the feature in object scale, focusing on intact land objects of building areas and parking spots. The hybrid structure shows a strong response to the obvious objects with both fine edges and intact areas. 

In general, the ablation results demonstrate two findings: \textbf{(1)} The PLAT module can stably optimize the framework training and reasonably exploit the LR labels during the large-scale HR land-cover mapping process. \textbf{(2)} The parallel CNN and Transformer branches are indispensable parts of the framework, which construct a more robust feature extractor to bridge local and global contextual information.
\section{Conclusion}
\vspace{-0.5em}

In this paper, a weakly supervised CNN-Transformer framework, Paraformer, is proposed to update large-scale HR land-cover maps in an HR-label-free, end-to-end manner. Experiments on two datasets show that Paraformer outperforms other approaches in guiding semantic segmentation of large-scale HR remote-sensing images with easy-access LR land-cover data. Further analysis reveals that the Paraformer can robustly adapt various landforms of wide-span areas and stably exploit different LR labels in producing accurate HR land-cover maps. The ablation studies demonstrate the effectiveness of the parallel CNN-Transformer structure and the PLAT module. Moreover, intermediate results of each training process and visualized contexts of each branch are demonstrated to transparently explain the components of Paraformer. In general, the proposed Paraformer has the potential to become an effective method for facilitating large-scale HR land-cover mapping.\vspace{+1em}
\\
\textbf{\large Acknowledgments}\vspace{+0.5em}
\\
This work has been supported by the National Key Research and Development Program of China (grant no. 2022YFB3903605) and the National Natural Science Foundation of China (grant no.42071322).

{
    \small
    \bibliographystyle{ieeenat_fullname}
    \bibliography{main}
}
\clearpage


\section*{Supplementary material}

# Supplementary Material - Learning without Exact Guidance: Updating Large-scale High-resolution Land Cover Maps from Low-resolution Historical Labels

In this supplementary, we provide a detailed description of the proposed framework and dataset organization. More experimental results are also presented. These three parts are demonstrated sequentially.

## A . Details of Paraformer

In the proposed Paraformer, a robust feature extractor parallel hybrids a downsampling-free CNN branch with a Transformer branch. To demonstrate the structures of CNN and Transformer branches more clearly, Figures S1 and S2 show the basic units of CNN and Transformer branches.

In this section, we focus on illustrating the basic units of the CNN branch in detail. The resolution preserving (RP) block shown in Figure S1 was firstly proposed in our previous work: L2HNet[1]. Here, we use $\mathbf{I}^{(b)}$, $\mathbf{M}^{(b)}$, and $\mathbf{F}^{(b)}$ to denote the input, middle, and fusion feature maps of the $b$-th block. Specifically, the input feature map of the first block is generated by a $3 \times 3$ convolution input layer with four input channels (i.e., the R-G-B-NIR bands of the images) and $C_I$ output channels. Therefore, the input feature map of the first block can be expressed as $\mathbf{I}^{(1)} \in \mathbb{R}^{N \times C_I \times H_I \times W_I}$, where $N$ represents the batch size and $C_I \times H_I \times W_I$ represents the channels, height, and width of the map, respectively. For the operation symbols, we represent a one-stride $(n \times n)$ convolutional layer with $C_1$ input channels and $C_2$ output channels as $W_{C_1,C_2}^{n \times n}$ (with padding when $n = 3, 5$). In addition, the batch normalization layer with the rectified linear unit (ReLU) function is simply denoted by $bn(\cdot)$, and $*$ represents the convolution operator. Based on this, the multi-scale feature fusion process from $\mathbf{I}^{(b)}$ to $\mathbf{M}^{(b)}$ can be described as:

$$\mathbf{M}^{(b)} = \text{concat} \begin{bmatrix} bn(\mathbf{I}^{(b)} * W_{C_I,C_I}^{1\times 1}), \\ bn(\mathbf{I}^{(b)} * W_{C_I,\frac{C_I}{2}}^{3\times 3}), \\ bn(\mathbf{I}^{(b)} * W_{C_I,\frac{C_I}{4}}^{5\times 5}) \end{bmatrix}. \quad (\text{S1})$$

As shown in Eq. (S1), the kernel numbers of the multi-scale convolutional layers are set to $\omega = \{\sqrt{2^{(1-n)}}\}_{n=1,3,5}$, which is inversely proportional to their kernel sizes.

Subsequently, we adopt a $1 \times 1$ convolutional layer after the concatenation of the multi-scale layers to reduce the dimensions of $\mathbf{M}^{(b)}$ from $C_I (1 + 1/2 + 1/4)$ to $C_I$, thus keeping the blocks lightweight. In addition, to maintain the shallow features and put residual learning into effect, a shortcut connection is adopted from $\mathbf{I}^{(b)}$ to $\mathbf{F}^{(b)}$. As a result, the final $\mathbf{F}^{(b)}$ can be described as:

$$\mathbf{F}^{(b)} = bn(\mathbf{M}^{(b)} * W_{C_I(1+1/2+1/4),C_I}^{1 \times 1}) + \mathbf{I}^{(b)}. \quad (\text{S2})$$

From Eqs. (S1)–(S2), $\mathbf{F}^{(b)}$ is a multi-scale fusion feature map with the same size, channels, and resolution as $\mathbf{I}^{(b)}$. Based on the structures, the RP block synchronously combines the multi-scale fusion attributes and residual learning ability to appropriately prevent the feature resolution reduction caused by the over-downsampling. Furthermore, after the feature fusion of several RP blocks, the predictions and corresponding CP maps are generated through a classifier that is constructed by a SoftMax function and a 1×1 convolutional layer $W_{C_I,L}^{1\times 1}$, where $C_I = 128$ is the channel numbers maintained in the entire backbone and $L$ is the output channel determined by the number of land-cover categories.

Moreover, the basic unit of the Transformer branch is shown in Figure S2, which includes a layer normalization (Layer Norm), multi-head self-attention (MSA), and multi-layer perception (MLP).

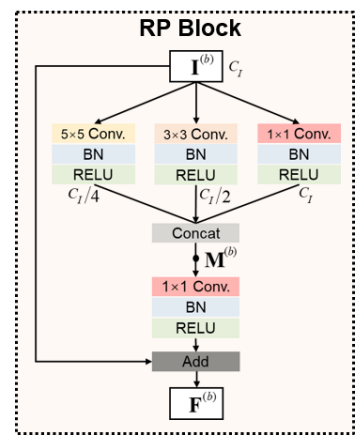

Figure S1. An illustration of an RP block. The input map $\mathbf{I}^{(b)}$ is sampled by three convolutional layers with sizes of $1 \times 1$, $3 \times 3$, and $5 \times 5$, and the convolution kernels in each layer are set to the proportion of $\omega$ for preventing feature resolution reduction caused by the over downsampling.

---
[1] https://doi.org/10.1016/j.isprsjprs.2022.08.008

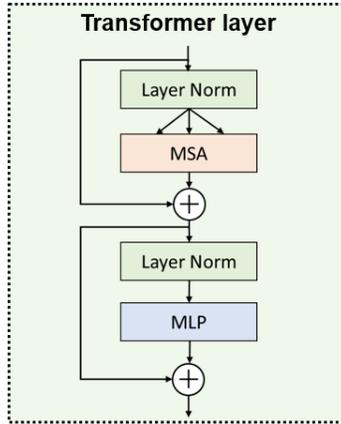

Figure S2. An illustration of a Transformer layer. The layer includes layer normalization (Layer Norm), multi-head self-attention (MSA), and multi-layer perception (MLP).

## B . Details of Study area and using data

In this section, we demonstrate the details of two large-scale datasets. Figures S3 and S4 show the location, coverage, and data samples of the Chesapeake Bay dataset and the Poland dataset. Tables S1 and S2 show the land-cover class unifying relations between the LR labels and HR ground truths.

**The Chesapeake Bay dataset:** The Chesapeake Bay, as the largest estuary in the USA, is about 320 kilometers long from its northern headwaters in the Susquehanna River to its outlet in the Atlantic Ocean. The Chesapeake Bay watershed covers about 160,000 $km^2$ areas of the surrounding drainage basin. It includes six administrative states of the USA which are New York, Pennsylvania, Delaware, Maryland, Virginia, and West Virginia. The Chesapeake Bay watershed contains various landforms with abundant ecological communities and diverse flora which brings challenges for large-scale high-resolution (HR) land-cover mapping. The Chesapeake Bay dataset, grouped by Microsoft[2], contains 1-meter resolution images and a 30-meter resolution land-cover product as the training data pairs and also contains a 1-meter resolution ground reference for assessment. Figure S3 illustrates the location, Digital Elevation Model (DEM), numbers of the tiles, and data samples of the Chesapeake Bay dataset. In more detail, the data sources are shown as follows:

1. The HR remote sensing images with 1-meter resolution were captured by the airborne platform of the U.S. Department of Agriculture's National Agriculture Imagery Program (NAIP). The images contained four bands of red, green, blue, and near-infrared.

2. The rough historical land-cover products with 30-meter resolution were collected from the National Land Cover Database of the United States Geological Survey (USGS). The NLCD data contains 16 land-cover types and is utilized as the labels during the training process of the proposed Paraformer framework.

3. The HR ground references with 1-meter resolution were obtained from the Chesapeake Bay Conservancy Land Cover (CCLC) project. The CCLC data were interpreted based on the 1-meter NAIP imagery and LiDAR data containing six land-cover types. In this paper, the CCLC data were only used as the ground reference for quantitative and qualitative assessment and were not involved in the framework training or optimization process.

**The Poland dataset:** The Republic of Poland has a territory traversing the Central European Plain and extends from Baltic Sea in the north to the Sudeten and Carpathian Mountains in the south. Topographically, with the flat, long sea lie and the hilly, mountainous terrain, the landscape of Poland is characterized by diverse landforms, river systems, and ecosystems. The Poland dataset contained 14 Provinces of Poland which included the Provinces of Pomorskie, Lódzkie, Lubuskie, Dolnoslaskie, and so on. The Poland dataset contains 0.25-meter resolution images, three kinds of 10-meter resolution land-cover products, and a 30-meter resolution land-cover product to construct the training data pairs with different combinations. Figure S4 demonstrated the location, DEM, numbers of the tiles, and data samples of the Poland dataset. In more detail, the data sources are shown as follows:

1. The HR remote sensing images with 0.25-meter and 0.5-meter resolution were collected from the Land-Cover.ai dataset where the image sources are from the public geodetic resource used in the Land Parcel Identification System (LPIS). The images contained three bands of red, green, and blue.

2. The rough historical labeled data with 10-meter resolution were collected from three types of global land-cover products which were (1) The FROM_GLC10 provided by the Tsinghua University, (2) The ESA_WorldCover v100 provided by the European Space Agency (ESA), and (3) The ESRI 10-meter global land cover (abbreviated as ESRI_GLC10) provided by the ESRI Inc. and IO Inc. The 30-meter resolution labeled data were collected from the 30-meter global land-cover product GLC_FCS30 provided by the Chinese Academy of Sciences (CAS).

3. The HR ground references were obtained from the OpenEarthMap dataset provided by the University of Tokyo. The ground references were interpreted based on the 0.25-meter and 0.5-meter resolution LPIS imagery and contained five land-cover types.

---
[2]https://lila.science/datasets/chesapeakelandcover

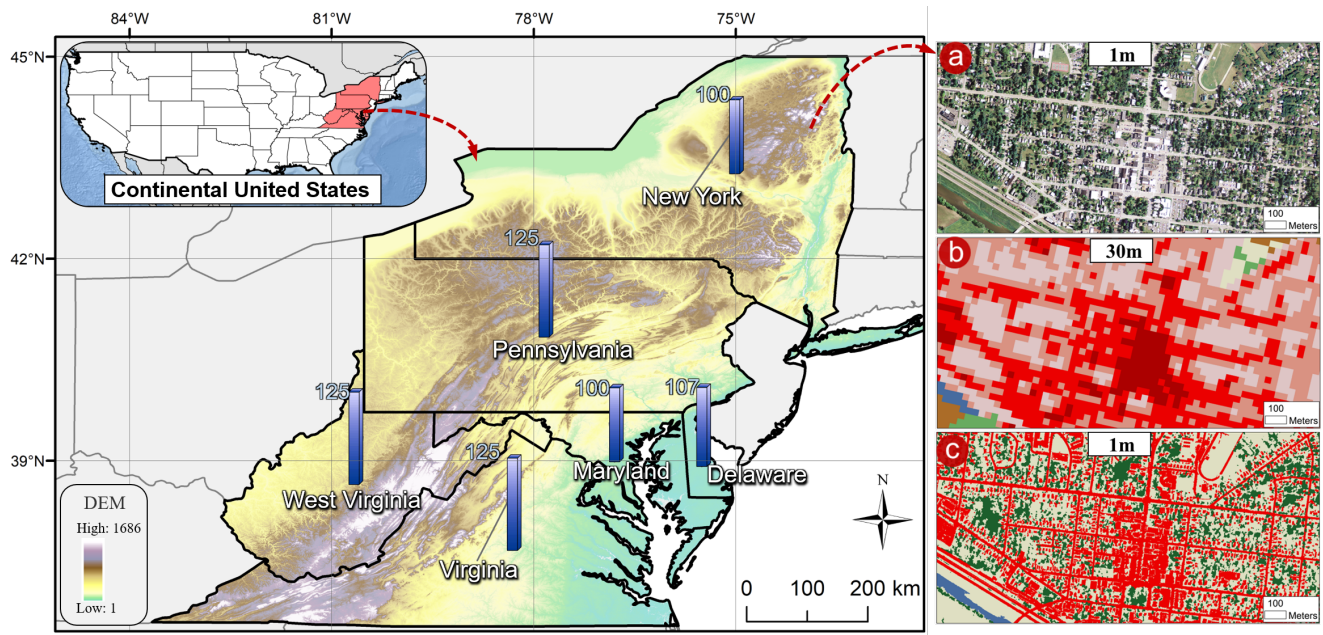

Figure S3. The Chesapeake Bay dataset covers six states of the USA, including the data sources of (a) The 1-m NAIP imagery, (b) The 30-m NLCD labels, and (c) The 1-m ground truth. The blue columns show the number of tiles.

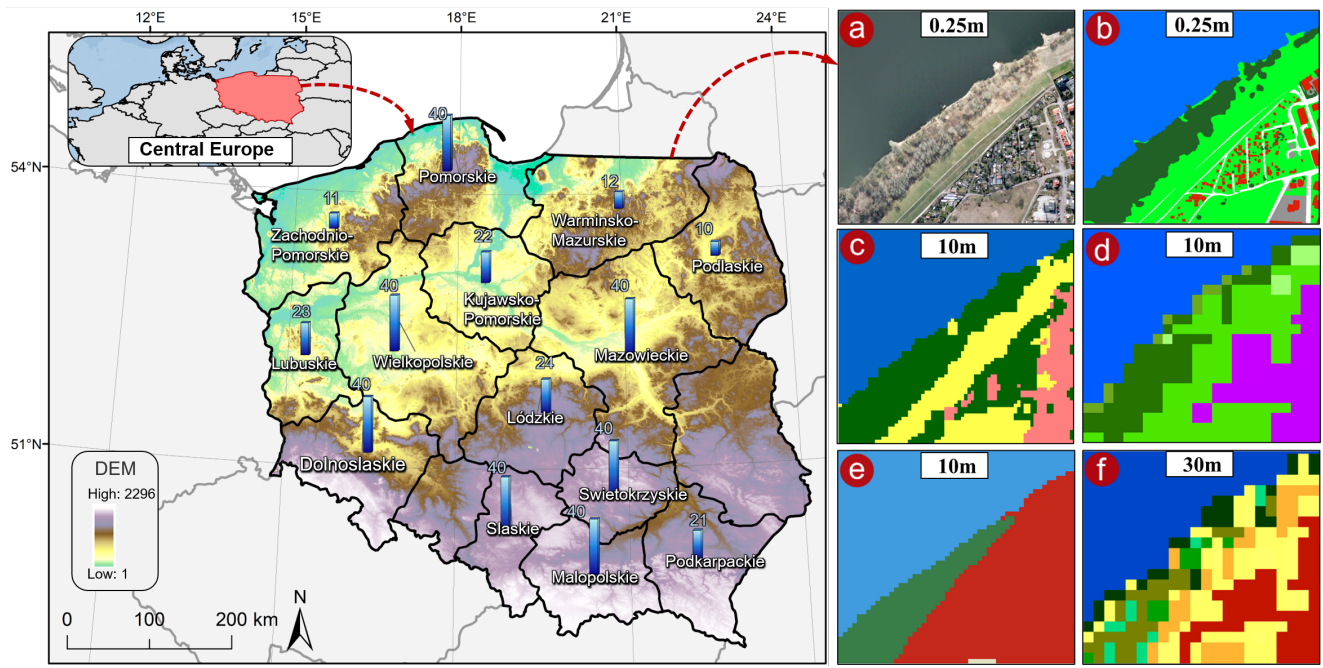

Figure S4. The Poland dataset covers 14 provinces of the country, including the data sources of (a) The 0.25-m imagery, (b) The 0.25-m ground truth, (c) The 10-m FROM_GLC10, (d) The 10-m ESA_GLC10, (e) The 10-m ESRI_GLC10, and (f) The 30-m GLC_FCS30. The blue columns show the number of tiles.

## C. Supplementary experiment results

To comprehensively demonstrate the performance of Paraformer, we sequentially illustrate supplementary experiment results as follows:

**Visual results of the Chesapeake Bay dataset:** Figures S5–S7 demonstrate one large-scale and two small-scale visual comparisons between Paraformer and four typical methods. From these visual results, the Paraformer is able to update accurate HR land-cover maps from the HR images source and LR label guidance. TransUNet shows clear

| Name | NLCD | CCLC | | |
|---|---|---|---|---|
| Affiliation | USGS, USA | Chesapeake Conservancy, USA | | **Target classes** |
| Resolution | 30 meters | 1 meter | | |
| Class | ▫ Developed open space<br>▫ Developed low c<br>▫ Developed medium<br>▫ Developed high | Roads<br>Building<br>Barren | | ▪ Built-up |
| | ▫ Deciduous forest<br>▫ Evergreen forest<br>▫ Mixed forest<br>▫ Woody wetland | Tree canopy | | ▪ Tree canopy |
| | ▫ Barren land<br>▫ Shrub/Scrub<br>▫ Grassland<br>▫ Pasture/Har<br>▫ Cultivated crops<br>▫ Herbaceous wetlands | Low vegetation | | ▪ Low vegetation |
| | ▫ Open water | Water | | ▪ Water |
| Note: | USGS= United States Geological Survey; | | | |

Table S1. Land-cover class unifying relations between the LR labels (NLCD) and HR ground truths. The first column shows the legends of LR labels. The last column shows the target classes for accuracy assessment and their colors shown in the visual results.

| Name | FROM_GLC10 | ESRI_GLC10 | ESA_GLC10 | GLC_FCS30 | OpenEarthMap | **Target classes** |
|---|---|---|---|---|---|---|
| Affiliation | THU, China | Esri&IO, USA | ESA, Europe | CAS, China | UTokyo, Japan | |
| Resolution | 10 meters | 10 meters | 10 meters | 30 meters | 0.25/0.5 meter | |
| Class | ▪ Forest | ▪ Trees | ▪ Trees | ▪ Deciduous broadleaved forest<br>▪ Open deciduous broadleaved forest<br>▪ Evergreen needle-leaved forest<br>▪ Mixed leaf forest | Tree | ▪ Tree canopy |
| | ▪ Shrubland<br>▪ Grassland<br>▪ Cropland | ▪ Scrub/Shrub<br>▪ Grassland<br>▪ Crops | ▪ Shrubland<br>▪ Grassland<br>▪ Cropland | ▪ Orchard<br>▪ Sparse shrubland<br>▪ Grassland<br>▪ Herbaceous cover<br>▪ Rainfed cropland<br>▪ Irrigated cropland | Rangeland<br><br>Agriculture land | ▪ Low vegetation |
| | ▪ Impervious area | ▪ Built Area | ▪ Built-up | ▪ Impervious surfaces | Building<br>Road<br>Developed space | ▪ Built-up |
| | ▪ Water body | ▪ Water | ▪ Open water | ▪ Water body | Water | ▪ Water |
| Note: | THU=Tsinghua University; ESRI=ESRI Inc.; IO=IO Inc.; ESA=European Space Agency; CAS=Chinese Academy of Science; UTokyo=The University of Tokyo | | | | | |

Table S2. Land-cover class unifying relations among four types of LR labels and HR ground truths. The 1–4 column shows the legends of LR labels. The last column shows the target classes for accuracy assessment and their colors shown in the visual results.

urban patterns but underestimates the built-up areas. UNet, as a typical CNN-based encoder-decoder framework, has a rough result consistent with the LR labels. L2HNet, as the state-of-the-art method for updating HR land-cover results from LR labels, shows an accurate edge of land objects but still has incorrect fragments in the results. RF, as a pixel-to-pixel learning method, has the finest edges but lacks of contextual information learning, which causes insufficient results overall (underestimating the water and low vegetation).

**Visual results of the Poland dataset:** Figures S8–S11 show the visual comparison between Paraformer and the other three typical methods which are trained with different LR land-cover labels. From the visual results, the Paraformer is able to refine a clear land-cover pattern from different types of LR land-cover labels. Even though some of the classes in the demonstration patches are not contained, Paraformer can jointly capture the local and global contexts and produce HR results that are consistent with the HR images.

**Further discussion:** In this part, we demonstrate more details of the loss fluctuation and supplementary large-scale experiments in China. Figure S12 shows the loss functions of $\mathcal{L}_{ce}$ and $\mathcal{L}_{mce}$ during framework training. The two training losses are stable to decrease in six states of the Chesapeake Bay dataset. This further indicates the robustness of the pseudo-label-assisted training (PLAT) module in learning from inexact LR labels. To further discuss the applicability of Paraformer, we conduct large-scale experiments in the whole of Wuhan City, China.

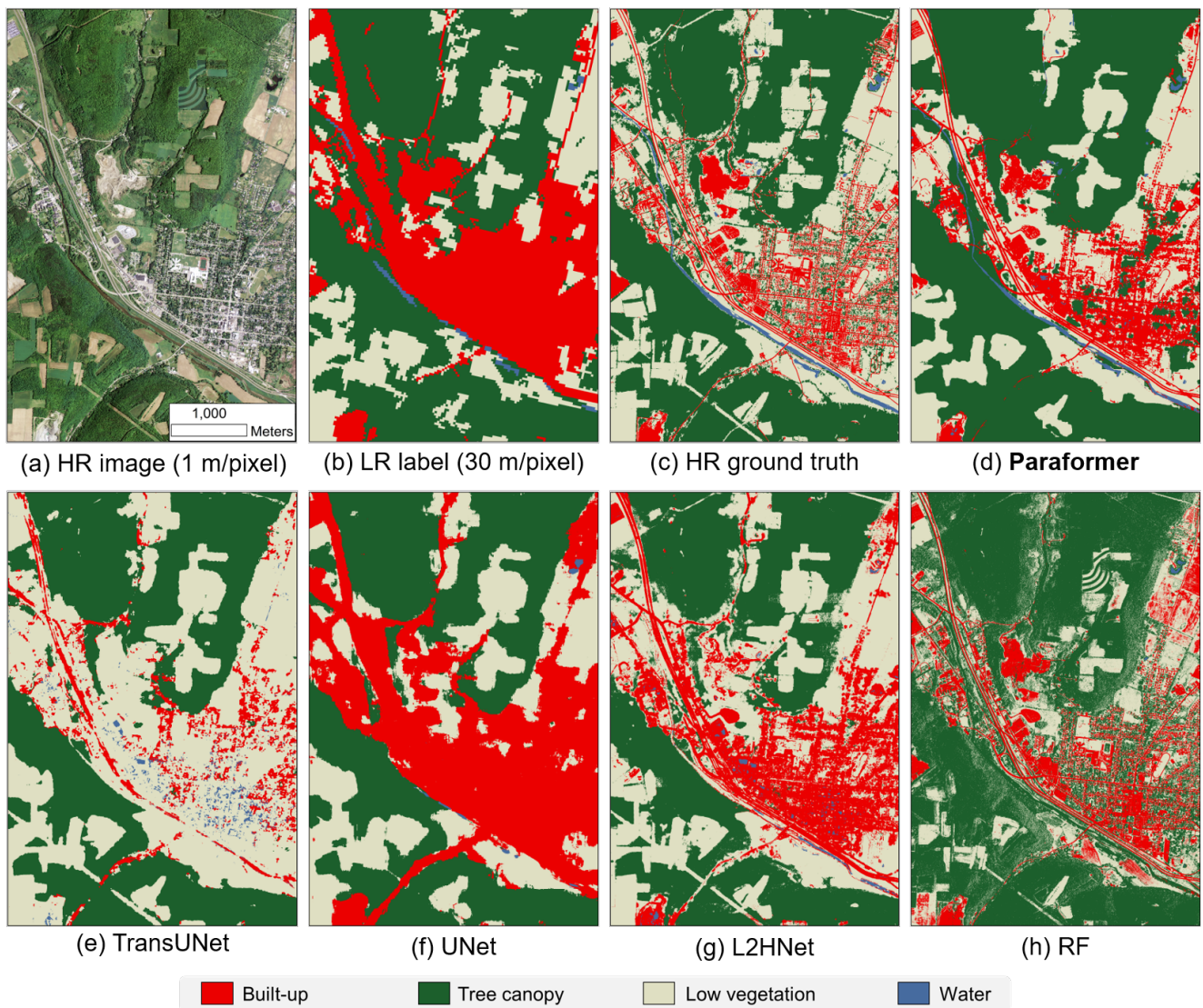

Figure S5. Demonstration of the training data and visual comparisons of the **Paraformer** and other typical methods on the Chesapeake Bay dataset with four unified classes. (a) HR image. (b) LR label. (c) HR ground truth. (d) land-cover mapping result of Parafomer. (e–h) land-cover mapping results of four typical methods.

Based on our previous work on SinoLC-1[3] (i.e., the first 1-m land-cover map of China), we regard the intersected results of three 10-m land-cover products (ESA_GLC10, Esri_GLC10, and FROM_GLC10) as the LR training labels of 1-m Google Earth images. As shown in Fig. S13 (a), the 1-m Google Earth image reveals clear land details. Fig. S13 (b–d) demonstrates three types of 10-m land-cover products. Compared with the original 1-m SinoLC-1 shown in Fig. S13 (e), the Paraformer is able to refine a more accurate urban pattern shown in Fig. S13 (f). For the whole of Wuhan City, the reported overall accuracy (OA) of SinoLC-1 is 72.40%. The updated results of the proposed Paraformer reach 74.98% with a 2.58% improvement.

---

[3] https://doi.org/10.5194/essd-15-4749-2023

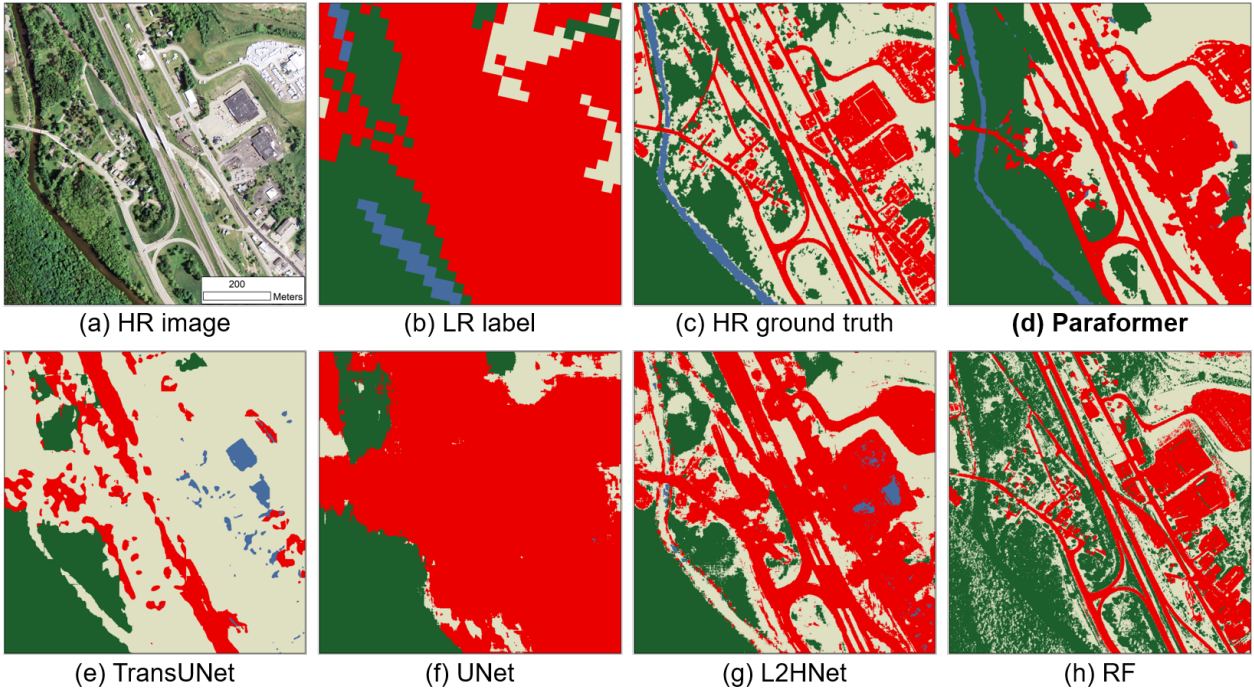

Figure S6. Sample A of the training data and visual comparisons of the **Paraformer** and other typical methods on the Chesapeake Bay dataset with four unified classes. (a) HR image. (b) LR label. (c) HR ground truth. (d) land-cover mapping result of Parafomer. (e–h) land-cover mapping results of four typical methods.

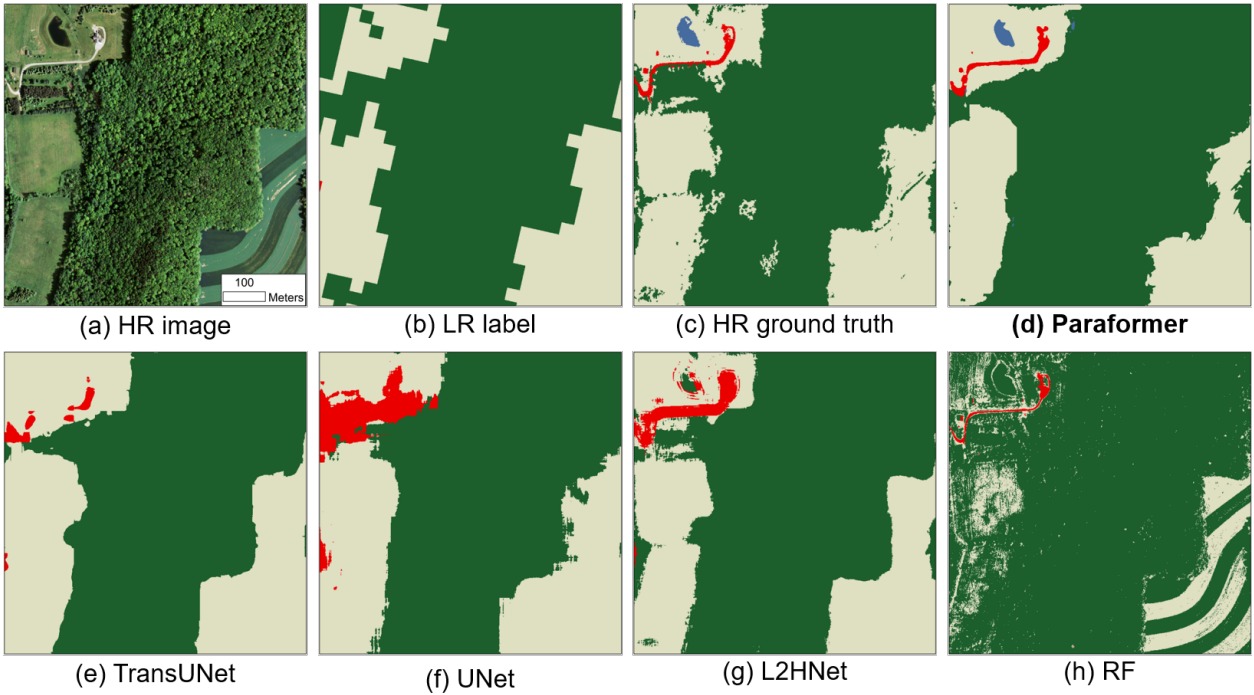

Figure S7. Sample B of the training data and visual comparisons of the **Paraformer** and other typical methods on the Chesapeake Bay dataset with four unified classes. (a) HR image. (b) LR label. (c) HR ground truth. (d) land-cover mapping result of Parafomer. (e–h) land-cover mapping results of four typical methods.

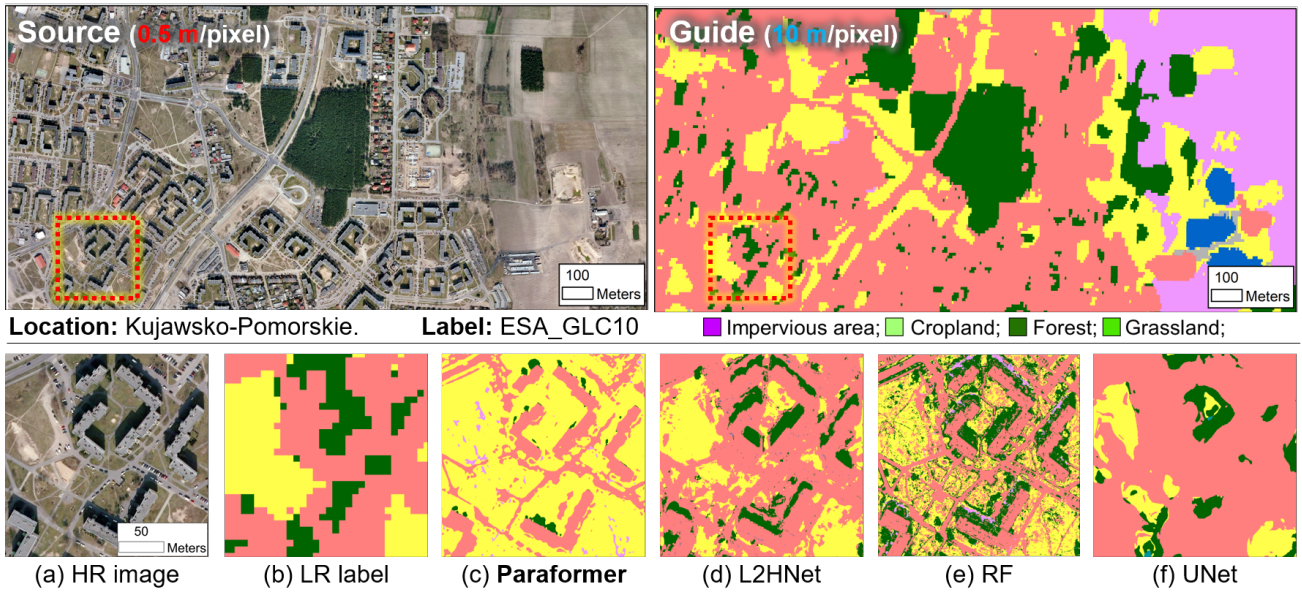

Figure S8. The visual results of Poland dataset with 10-m ESA_GLC10 training labels. (a) The 0.5-m image, (b) The 10-m label sampled from the ESA_GLC10. (c) Result of Paraformer. (d) Result of L2HNet. (e) Result of RF. (f) Result of UNet.

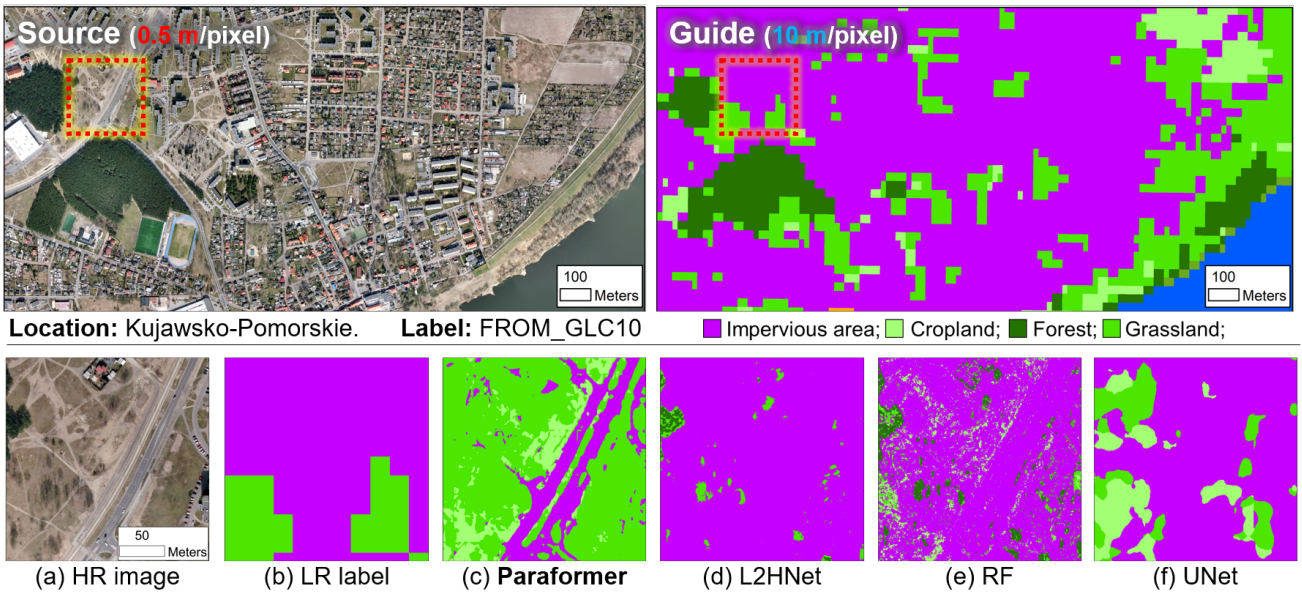

Figure S9. The visual results of Poland dataset with 10-m FROM_GLC10 training labels. (a) The 0.5-m image, (b) The 10-m label sampled from the FROM_GLC10. (c) Result of Paraformer. (d) Result of L2HNet. (e) Result of RF. (f) Result of UNet.

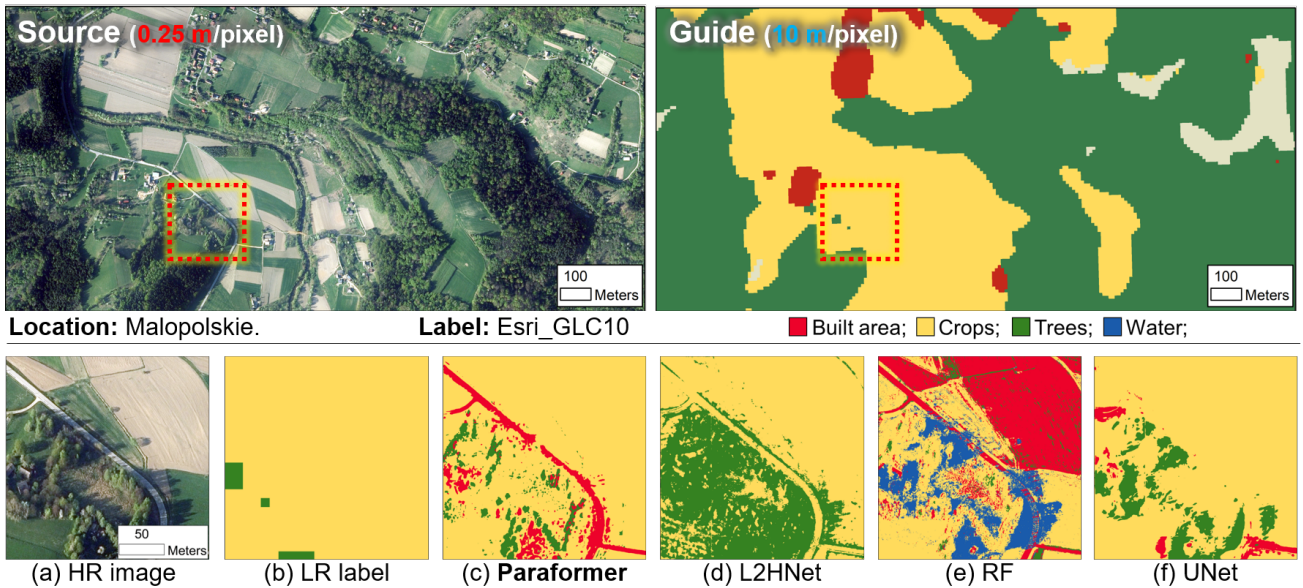

Figure S10. The visual results of Poland dataset with 10-m Esri_GLC10 training labels. (a) The 0.25-m image, (b) The 10-m label sampled from the Esri_GLC10. (c) Result of Paraformer. (d) Result of L2HNet. (e) Result of RF. (f) Result of UNet.

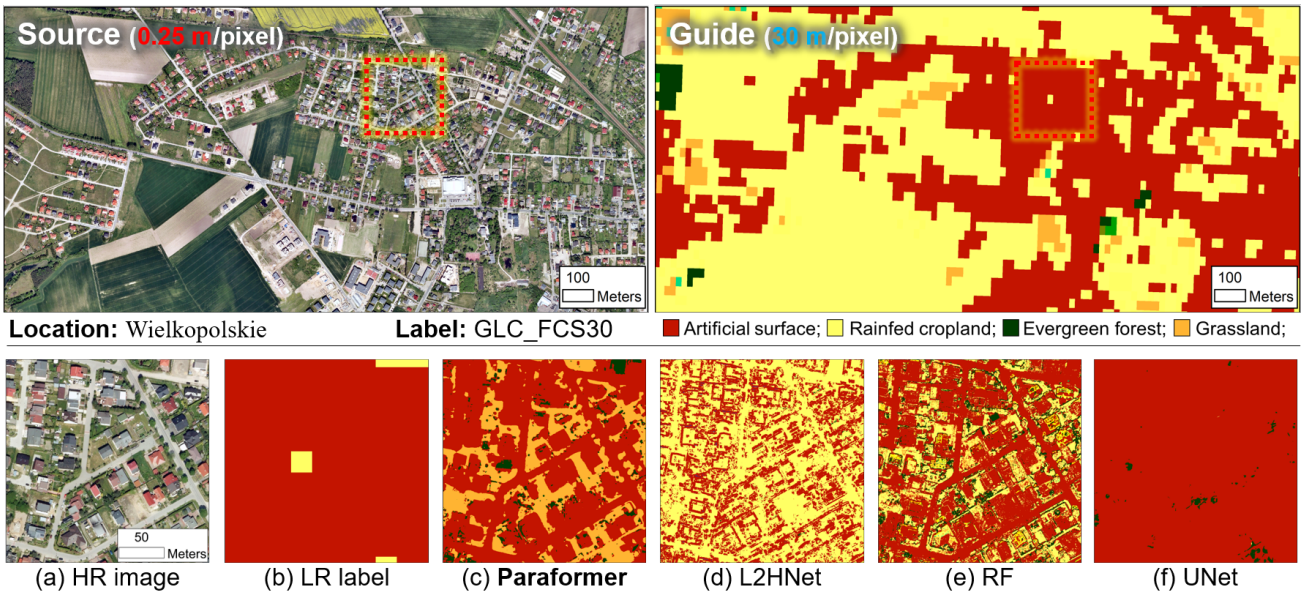

Figure S11. The visual results of Poland dataset with 30-m GLC_FCS30 training labels. (a) The 0.5-m image, (b) The 10-m label sampled from the GLC_FCS30. (c) Result of Paraformer. (d) Result of L2HNet. (e) Result of RF. (f) Result of UNet.

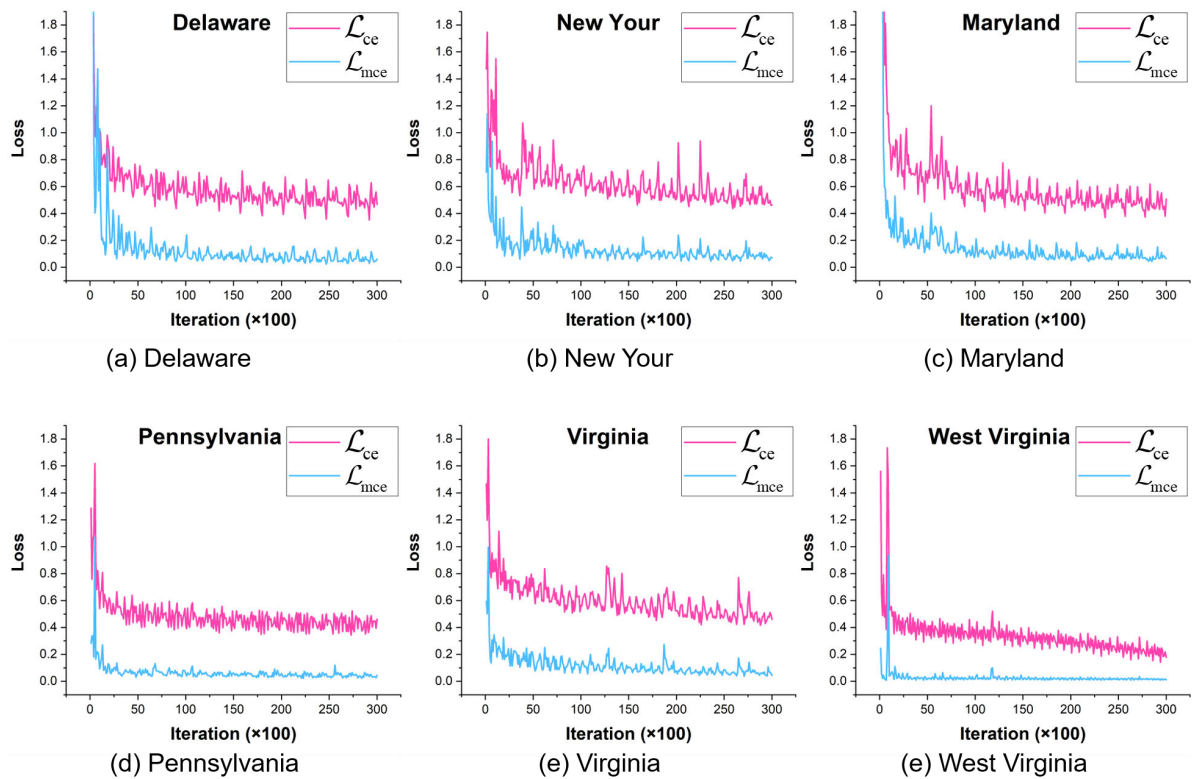

Figure S12. Demonstration of the loss functions $\mathcal{L}_{\text{ce}}$ and $\mathcal{L}_{\text{mce}}$ during framework training. Sub-figures (a)–(e) demonstrate the training process in six states of the Chesapeake Bay dataset.

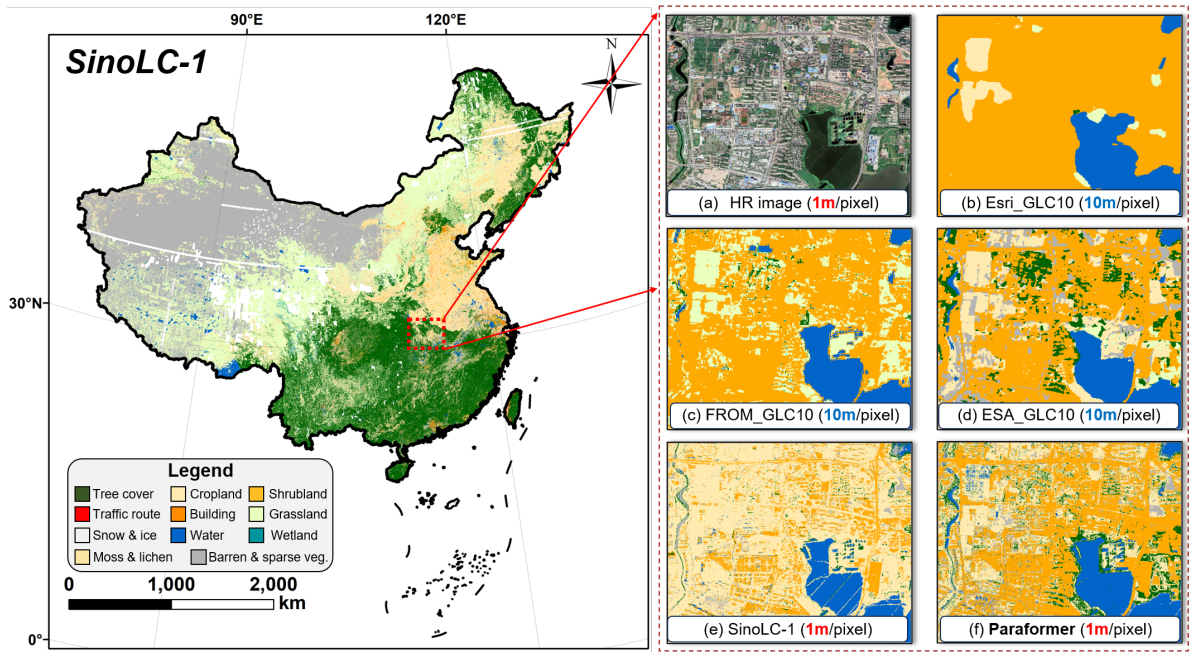

Figure S13. Demonstration of the supplementary experiments of SinoLC-1 dataset. The visual results are sampled from Wuhan, China. The Paraformer is used to update the 1-m land-cover map in the whole of Wuhan City, reporting an OA of 74.98%.



\end{document}